\documentclass[letterpaper, 10 pt, conference]{ieeeconf}  % Comment this line out if you need a4paper

\IEEEoverridecommandlockouts                              % This command is only needed if 
\usepackage{svg}
\usepackage{amsmath}
\usepackage{subcaption}
\usepackage{times}
\usepackage{amssymb,url}
\usepackage{algorithmic,algorithm}
\usepackage{multirow}
\usepackage{mathabx}
\usepackage{graphicx}
\usepackage{color}
\usepackage[export]{adjustbox}
\usepackage{multicol}
\usepackage[bookmarks=true]{hyperref}
\usepackage{cancel}
\usepackage{lipsum}

\newtheorem{theorem}{Theorem}[section]
\newtheorem{corollary}[theorem]{Corollary}

\newtheorem{definition}{Definition}[section]

\newcommand\itemf{\item[$F$:]}
\newcommand\itemr{\item[$R$:]}

\newcommand\itemq{\item[$Q$:]}
\newcommand\itemt{\item[$T$:]}

\newcommand\itemi{\item[$I$:]}
\newcommand\itemd{\item[${\Delta}$:]}

\newcommand\blfootnote[1]{%
  \begingroup
  \renewcommand\thefootnote{}\footnote{#1}%
  \addtocounter{footnote}{-1}%
  \endgroup
}

\begin{document}
\title{MTMR Assignment}

% You will get a Paper-ID when submitting a pdf file to the conference system
% \title{\LARGE \bf
% Making Assignments for Multitasking Robots in Multi-Robot Tasks
% }

% why emphasizing multi-robot tasks?
% Your approach would also work for single-robot tasks, right?
\title{\LARGE \bf
Assigning Multi-Robot Tasks to Multitasking Robots
}

% \author{Albert Author$^{1}$ and Bernard D. Researcher$^{2}$% <-this % stops a space
% \thanks{*This work was not supported by any organization}% <-this % stops a space
% \thanks{$^{1}$Albert Author is with Faculty of Electrical Engineering, Mathematics and Computer Science,
%         University of Twente, 7500 AE Enschede, The Netherlands
%         {\tt\small albert.author@papercept.net}}%
% \thanks{$^{2}$Bernard D. Researcheris with the Department of Electrical Engineering, Wright State University,
%         Dayton, OH 45435, USA
%         {\tt\small b.d.researcher@ieee.org}}%
% }

\author{Winston Smith and Yu Zhang}% <-this % stops a space
%\thanks{*This work was not supported by any organization}% <-this % stops a space

%\author{\authorblockN{Michael Shell}
%\authorblockA{School of Electrical and\\Computer Engineering\\
%Georgia Institute of Technology\\
%Atlanta, Georgia 30332--0250\\
%Email: mshell@ece.gatech.edu}
%\and
%\authorblockA{Twentieth Century Fox\\
%\and
%\authorblockN{James Kirk\\ and Montgomery Scott}
%\authorblockA{Starfleet Academy\\
%San Francisco, California 96678-2391\\

% avoiding spaces at the end of the author lines is not a problem with
% conference papers because we don't use \thanks or \IEEEmembership

%Homer Simpson\authorrefmark{2},
%Montgomery Scott\authorrefmark{3} and
%Eldon Tyrell\authorrefmark{4}}
%Telephone: (800) 555--1212, Fax: (888) 555--1212}
%\authorblockA{\authorrefmark{4}Tyrell Inc., 123 Replicant Street, Los Angeles, California 90210--4321}}

\maketitle
\thispagestyle{empty}
\pagestyle{empty}

%%%%%%%%%%%%%%%%%%%%%%%%%%%%%%%%%%%%%%%%%%%%%%%%%%%%%%%%%%%%%%%%%%%%%%%%%%%%%%%%
\begin{abstract}

One simplifying assumption in existing and well-performing task allocation methods is that the robots are single-tasking: each robot operates on a single task at any given time. While this assumption is  
harmless
to make in some situations,  
it can be inefficient 
or even infeasible in others. 
In this paper, we consider assigning multi-robot tasks to multitasking robots.
The key contribution is a novel task allocation
framework that incorporates the consideration of physical constraints introduced by multitasking. 
This is in contrast to the existing work where such constraints are largely ignored. 
After formulating the problem, we propose 
a compilation 
to weighted MAX-SAT, which allows us to leverage existing solvers for a solution. 
A more efficient greedy heuristic is then introduced. 
For evaluation, we first compare our methods with a modern baseline that is efficient for single-tasking robots to validate the benefits of  multitasking in synthetic domains. 
Then, using a site-clearing scenario in simulation, we further
 illustrate the complex task interaction considered by the multitasking robots in our approach to demonstrate its performance. 
%Finally, we demonstrate an experiment with physical robots to show a scenario where multitasking requiring interleaving of tasks is preferable to singletasking with multiple robots.
Finally, we demonstrate a higher-complexity simulation to demonstrate the scalability and applicability of our approach.

\vskip-5pt

\end{abstract}

%%%%%%%%%%%%%%%%%%%%%%%%%%%%%%%%%%%%%%%%%%%%%%%%%%%%%%%%%%%%%%%%%%%%%%%%%%%%%%%%
\section{Introduction}
\blfootnote{The authors are with the School of Computing and Augmented Intelligence,
        Arizona State University
        {\tt\small \{wtsmith7, yzhan442\}@asu.edu}.}%
In prior work on task allocation, it has been a near-omnipresent assumption that robots are single-tasking~\cite{korsah2013comprehensive, gerkey}. 
When considering 
task allocation  
practically, it is often more convenient to have robots work on multiple tasks simultaneously 
to improve efficiency,
and may in fact be required in some situations.
For example, 
consider a scenario where a robot needs to swipe a keycard to unlock a door temporarily and also open that door at the same time. 
We note that it is not always possible for multiple robots to ``divide and conquer'' these tasks due to a space constraint; under certain conditions (e.g., robots with large footprints), one robot \textit{must} do both tasks with two manipulators since two robots cannot fit in the space around the door simultaneously.
The ability for a multi-robot system to automatically identify this infeasibility so that a bi-manual robot can be assigned instead of two robots is missing.
Redesigning this scenario such that there is a single ``swipe keycard and open door'' task to force the task to be assigned to a single robot is possible, but
this approach effectively requires the multitasking solution to already be known so that we can combine each set of tasks that require multitasking into a single task.
A generic solution for multi-robot tasks with multi-tasking robots requires the automatic decomposition and combination of tasks, which goes beyond simply adding up or splitting task requirements and is even more complicated.
Hence, compiling such problems to problems with single-tasking robots is utterly impractical in general.
%{\color{blue} Note that this disqualifies compilation to the domain considering only single-tasking robots.}

The key observation here is that achieving multitasking robots with multi-robot tasks is not only about the ability of each robot to perform its tasks or parts of the tasks assigned to it (in the case of multi-robot tasks) in terms of having appropriate sensors, motors, etc., 
but also about the ability of each robot to perform its assignments in terms of \textit{physicality} --
that is, whether these assignments introduce physical constraints that prevent the robot from achieving them simultaneously (in the case of multitasking) or hinder robots from cooperating on a task (in the case of multi-robot tasks).
In the previous example, 
the space constraint, a type of physical constraint, introduced the required multitasking.
At the same time, multitasking can introduce other physical constraints. 
Following from above, 
to be able open the door, a bi-manual robot must be flexible enough to use one arm for swiping the keycard and another arm for opening the door simultaneously, which imposes a constraint on the arm span.

Hence, the key to enable multitasking with multi-robot tasks lies in the ability to identify and exploit synergies among these physical constraints to ensure their compatibility. 
A significant challenge is that constraints can be inter-dependent, meaning that they may interact with others in complex ways, such as implying other constraints recursively.
For example, if the robot has a fixed camera for vision, a constraint on the robot's viewing angle as a result of the keycard task would constrain the robot's facing and one axis of its position.
This complexity makes it substantially more challenging to anticipate the influence of a task assignment on future assignments when multitasking is considered with multi-robot tasks, not to mention the significantly grown number of candidate assignments. 
Prior works on multitasking robots with multi-robot tasks~\cite{shehory1998methods, shiroma2009comutar} ignore the inter-dependencies among constraints and hence can introduce invalid solutions. 
% While we may assume the knowledge of such %rules underlying these 
% implications \textit{a priori}, 
% the constraints and assignments are now interdependent.
% This is in contrast to the existing work where similar constraints are considered static~\cite{vig, shiroma2009comutar}.
% A naive approach would be to simply check every possible assignment of tasks to robots. 
% With multitasking robots, however, the number of such assignments is significantly larger than that with single-tasking robots, which makes this impractical except for the smallest problems. 
% If we assume single-tasking robots and single-robot tasks, the number of possible assignments for $10$ tasks and $10$ robots is $10^{10}$.
% In our problem, with multi-tasking robots and multi-robot tasks, this explodes to potentially $(10!)^{10!}$ possible assignments.

To the best of our knowledge, this work represents the first framework for assigning multi-robot (MR) tasks to multitasking (MT) robots with instantaneous assignment (IA) (MT-MR-IA in \cite{gerkey}) while considering physical constraints. 
It is a significant extension of prior work~\cite{smith} for determining task feasibility under a given set of task assignments to robots.
Since the mechanism for considering the physical constraints with multitasking robots and multi-robot tasks are the same, we focus on the more unique multitasking capability in the following discussion.
We present a general problem definition, explore a compilation approach to the weighted MAX-SAT problem first, and then a greedy method. 
Simulation results in synthetic domains demonstrate that our approach can handle medium-size problems effectively. 
In particular, we show that a modern baseline for task allocation with single-tasking robots performs poorly compared with our method, verifying the advantage of multitasking.  
%Even the simple greedy method (STAMR-G) outperforms the baseline significantly.
Using a site clearing scenario in simulation, we further illustrate the complex task interactions between multitasking robots that can be handled by our approach to demonstrate its capability. 
Finally, we show how our approach leads to increased task efficiency in a complex setting with realistic simulations.

\vskip-5pt

\label{sec:introduction} 

\section{Related Work}

The multi-robot task allocation (MRTA) problem has been a subject of prior research, reviewed in~\cite{gerkey, korsah2013comprehensive}.
% In particular, many prior works have focused on single-tasking robots with single-robot or multi-robot tasks (ST-SR/MR).
For multi-robot systems, one class of problems of particular interest involves multi-robot tasks (MR). 
Since even the simplest subclass that addresses single-tasking robots (ST) and instantaneous assignments (IA) is NP-hard, the focus has been on approximate solutions or heuristic methods~\cite{shehory, vig, sandholm, adams2011coalition, liemhetcharat2014weighted, gerkey2, parker2, zlot, zhang2013considering}.
%We take \cite{zhang2013considering}, the most performant of these, as a baseline for our evaluation.
The more general subclass with multitasking robots (MT) 
has not seen nearly as much examination, with some notable exceptions~\cite{shehory1998methods, shiroma2009comutar, yu2, yu3, parker2}.
Several recent works also approach the MT-MR domains.
For example, \cite{miloradovic} solves the MT-MR-TA problem as posed in \cite{gerkey}, and \cite{bischoff} considers precedence constraints between tasks.
Hence, they are not applicable to the MT-MR-IA setting we considered.
%{\color{red}If none of these more recent work were able to account for physical constraints, shouldn't they be merged into the first paragraph in related work, instead of being separately placed here?}
%{\color{blue} Good point; moved that text and this conversation.}
Furthermore, none of these prior works are general enough under the presence of physical constraints, which are in addition to resource constraints to fulfill task requirements and can introduce complex task dependencies.
%In our prior work~\cite{smith}, we studied the feasibility of multitasking under physical constraints given the set of task assignments to robots and we extend it here to create and optimize such assignments. %{\color{blue} Maybe "create" instead of "optimize"?}

%Physical constraints may assume various types,
Physical constraints may assume various forms, and are especially prevalent with multi-robot tasks and multitasking robots. 
Co-location constraints on sensor and motor capabilities, 
and spatial constraints due to physical presence, are examples of physical constraints. 
Most prior works on task allocation (especially with multi-robot tasks) only implicitly consider  physical constraints and thus have limited applicability.
For example, previous works have considered overlapping coalitions~\cite{shehory1996formation, dang2006overlapping} but assume the influences of these constraints are captured by the utilities of coalitions. 
That approach not only imposes a challenge on the domain designer but also implies that the constraints are independent across assignments, which is restrictive.

Co-location constraints have been explicitly considered in~\cite{vig} and generalized in~\cite{yu2, yu3, parker2}.
However, one common limitation of these prior works is that they all ignore the inter-dependencies between these constraints,
which can result in infeasible solutions. 
%Most prior works on MT-MR problem also consider such physical constraints, but in simplified forms and thus have limited applicability.
For multitasking, in particular, its feasibility depends critically on the compatibility of these constraints in complex and interactive ways, which are required to handle multiple tasks simultaneously.
Although multitasking robots have been considered before, such as in~\cite{shiroma2009comutar}, no physical constraints were considered.
A prior work~\cite{smith} that looked into generally formulating physical constraints, however, operates under a restricted problem setting where the set of assignments is given, which we will  extend in this work to create and optimize assignments.
A general approach to considering physical constraints would enable a multi-robot system to express real-world problems and generalize to novel scenarios more effectively.

The way in which physical constraints are formulated in this work resembles how works in action languages \cite{actionlanguages, khandelwal-bc, pddl+, pddl+2} deal with indirect action effects in planning, but our work examines these effects in robot task allocation:
 these prior works only examine single robots, whereas ours allows for multiple agents.
 ~\cite{alferes-lups} and \cite{qiao-eca} applied a similar concept for event modeling, which is applied to logic programs.
In addition, the indirect action effects in these prior works are generally not allowed to interact with each other.

%One approach commonly used when solving combinatorial problems is that of compilation to the SAT problem. An example that is well known is Boolean Satisfiability Based Planning (SATPLAN)~\cite{satplan}, which first introduced the idea of planning as satisfiability. 
%An advantage of such an approach is that many existing efficient SAT solvers can be leveraged.
%However, we have yet to see the application of similar ideas to complex task allocation problems.  
%Of particular relevance to this paper is the work in solving the Weighted MAXSAT variant, especially modern solvers such as in \cite{uwrmaxsat}; solvers for this variant of the problem can be applied directly to our solution method, as discussed later.

%%%%%%%%%%%%%%%%%%%%%%%%%%%%%%%%%%%%%%%%%%%%%%%%%%%%%%%%%%%%%%%%%%%%%%%%%%%%%%%%
\section{Approach}
\label{sec:overview}

%%%%%%%%%%%%%%%%%%%%%%%%%%%%%%%%%%%%%%%%%%%%%%%%%%%%%%%%%%%%%%%%%%%%%%%%%%%%%%%%
\subsection{Running Example}
\label{sec:running}

\begin{figure}
    \centering
    \includegraphics[width=0.7\columnwidth, height=0.37\columnwidth]{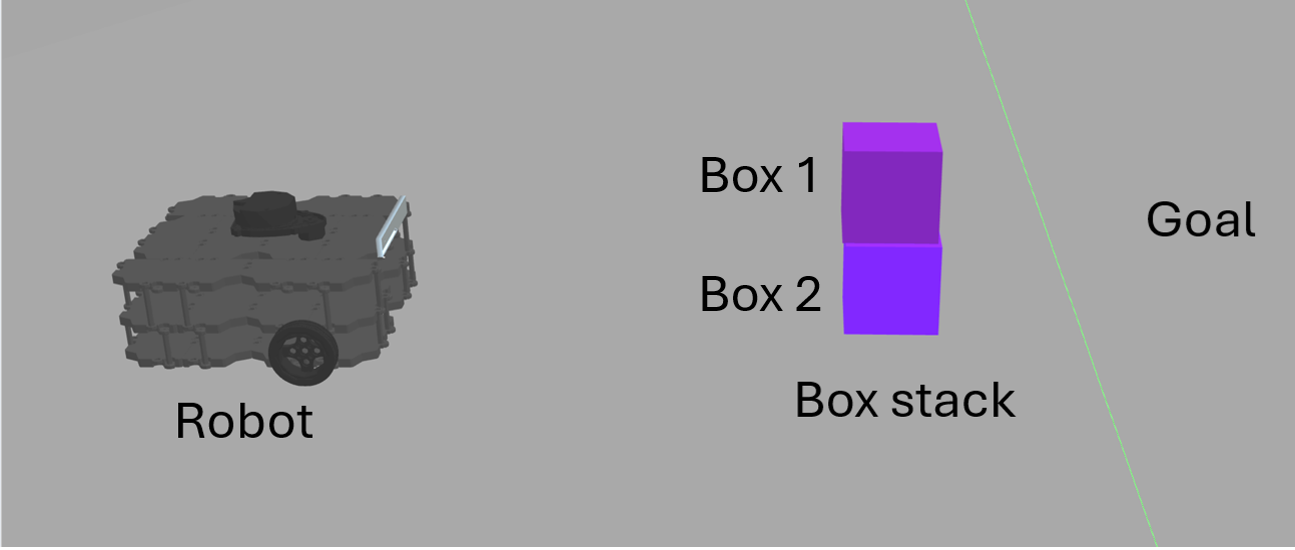}
    \vskip-3pt
    \caption{
    Running example in ROS with the boxes stacked. }
    \vskip-15pt
    \label{fig:motivating}
\end{figure}

As a running example, we will examine a scenario in which a robot must push several boxes to a goal area (see Fig. ~\ref{fig:motivating}); each box corresponds to a single task.
The two boxes are initially on the ground. 
Consider a situation where the robot only has enough 
battery power for one trip, or where the robot must push the boxes through a one-way door. 
In such a case, assuming the robot is capable of pushing both boxes at the same time, the robot is required to perform multitasking by stacking one box on top of the other and pushing them together by pushing the bottom box.
While this may be intuitive from a human's perspective, 
it involves substantial challenges that are representative of both the synergistic and restrictive aspects in multitasking. 
The robot must:

\begin{enumerate}
    \item (\textbf{\textit{Synergistic}}) Understand that multiple tasks can be achieved simultaneously, e.g., when pushing the bottom box, the top box is also pushed if they are stacked;
    \item (\textbf{\textit{Restrictive}}) Identify and respect any additional constraints as a result of multitasking, e.g., having a box on top of the bottom box effectively increases the weight of the bottom box and the pushing power required.
\end{enumerate}

With existing approaches, considering these aspects would require building the conditions of each aspect above into problem specifications on a problem-by-problem basis: e.g., combining the two tasks by redefining the box stack to be a single task with an updated weight.
This becomes infeasible when there are more boxes and robots, boxes have different weights,  heavy boxes should not be stacked on lighter ones, and robots have different pushing powers.
In this paper, we introduce a formalism that represents such knowledge in a \textit{problem-independent} way.
As a result, our formalism and solution methods together allow robots to flexibly reason about and solve novel scenarios {\color{black}in the same domain.}
%such as when there are more than two boxes to be moved, boxes have different weights, and robots have different pushing powers.

We assume that this knowledge is provided by users and leave learning it to future work.
%We also assume that the initial state is conducive to multitasking 
To focus on the task allocation challenge, 
we assume that the initial state of any given problem can be converted to a set of candidate initial states that are conducive to instantaneous assignment (IA).
For example, Fig. \ref{fig:motivating} could be a candidate initial state for the problem initial state where both boxes are on the ground. 
%In the running example, this means that, if the boxes are initially on the ground, a candidate initial state is when the two boxes are stacked;
This means that, before task execution, a preprocessing step must exist that can stack the boxes after task allocation is made.
%exists that would stack them  before task allocation.
%Updating the initial state such that synergies can be exploited to 
These steps require the consideration of complementary tasks as well as time-extended assignment. 
Thus, it is more natural to address them in future work when we  extend our formulation to time-extended domains.

% Prior work \cite{smith} introduced a method for considering only the constraining aspect above to determine the feasibility of a given set of task assignments to multitasking robots under the induced constraints.
% However, that work does not examine \textit{how} to make such assignments, only how to evaluate an assignment once it has been made.

We build our work on \cite{smith} to model physical constraints, even though only the restrictive aspect above was considered there. 
Our contributions are a) extending the formalism from \cite{smith} to consider synergies so that robots can reason about {\it how} to perform multiple tasks simultaneously (instead of being told how), b) providing the first general solution method for the MT-MR-IA problem considering physical constraints, and c) evaluating the effectiveness of multitasking in synthetic and simulated domains.

\subsection{Preliminaries \& Prior Work}

A predicate in logic is a boolean 
function that represents a property or relation \cite{huth}.
It may evaluate to true or false depending on its arguments.
In the
planning community, predicates are used to specify states or partial states.
In the running example, the predicate $F_{On}(o_1, o_2)$ is used to indicate whether or not $o_1$ is on top of $o_2$. $o_1$ and $o_2$ are both arguments to the predicate $F_{On}$, also called {\it referents}, which can remain uninstantiated, {indicated by capital letters}, 
or instantiated to domain elements, 
%A referent of a predicate may be instantiated to a specific domain element, 
indicated by lowercase letters with numeric subscripts, 
which are mapped to objects in the environment. 
In Fig. \ref{fig:motivating}, $F_{On}(o_1, o_2)$ is true when $o_1$ is mapped to Box 1 and $o_2$ to Box 2.

A physical constraint is a constraint on part of the physical configuration of the world.
We conservatively assume that a single physical constraint 
may arbitrarily restrict the \textit{physical configuration} that {satisfies the constraint.} %corresponds to the associated predicate being true.
For example, a constraint on $F_{On}(o_1, o_2)$ restricts $o_1$ to be on $o_2$.
%(for example, the exact location of $o_1$ on $o_2$)
% We also assume that the values of constraints may change arbitrarily over time.
The exact physical configuration that satisfies $o_1$ on $o_2$, however, may vary arbitrarily over time (i.e., where $o_1$ is on $o_2$ exactly).
As a result, we do not allow two physical constraints to apply to the same predicate since they could require incompatible physical configurations to be assumed.
For example, two physical constraints on the location of a box could require the box to be placed at different locations at the same time.
Such a stipulation is {\it conservative} since two constraints on the same predicate may still be compatible, e.g., when both constraints require the box to always be at exactly the same location or be at different locations at different times.
%and future work will extend the formulation to remove this limitation.

 %removed the next sentence
% If all of the referents in a predicate are instantiated as in $F_{Pos}(r_1)$, we refer to it as fully instantiated. 

It may be the case that a set of constraints implies additional constraints that are not in the original set.
{These %implied constraints 
must also be considered.}
For example, a constraint on $F_{Pos}(o_2)$ (i.e., the position of $o_2$) and another on $F_{On}(o_1,o_2)$ together imply a constraint on $F_{Pos}(o_1)$.
To capture this, {
we use CIRs, which are analogous to %that of 
inference rules} from~\cite{smith}:

\begin{definition}[Constraint Implication Rule (CIR)]
    Given a set of predicates $S$ and a predicate $f$, 
    a constraint implication rule specifies a relationship where the constraints on all of $S$ imply a constraint on $f$, written as $S \rightarrow f$.
    \label{def:infer-iis} 
\end{definition}

% In case of conflicts between CIRs, i.e., one CIR negates the precondition of another, precedence is decided by {\color{blue}a pre-specified order}. 
%order of definition.

For example, the example CIR described above could be written in its uninstantiated form as:
\begin{center}
\vskip-3pt
$\{F_{Pos}(Y), F_{On}(X,Y)\} \rightarrow F_{Pos}(X)$
\end{center}
\vskip-3pt
%which indicates that if $F_{Pos}(o_2)$ and $F_{On}(o_1,o_2)$ are constrained, then $F_{Pos}(o_1)$ would also be constrained.
Note that the same (different) uninstantiated referent label must be instantiated to the same (different) domain element in a CIR, e.g., $Y$ must instantiate to the same domain element for a given application of the CIR above.
CIRs may be given as uninstantiated or partially instantiated, but all referents must be instantiated when applying a CIR.
A CIR with uninstantiated or partially instantiated predicates thus specifies a set of instantiated CIRs.
We require any referents to $f$ to appear in $S$ - that is, any set of constraints cannot imply a constraint on new domain elements that are not referenced in that set of constraints.
A CIR is automatically triggered whenever constraints corresponding to its left hand side are all present. %Conflicts between CIRs, i.e., one CIR negates the precondition of another, are decided by a pre-specified order of CIRs.  
%(the next paragraph break was also added) 
When specifying CIRs, it is important to consider the physical processes behind how the environment acts upon itself, and to describe the effects of these processes in a general way as opposed to being case-specific.

{\color{black} CIRs are activated in the order they were defined, to prevent conflicts.}
We will show later how CIRs allow us to consider both the synergistic and restrictive aspects for multitasking.
% CIRs enable synergy as well as adding constraints, and in fact allow us to achieve tasks that require predicates which cannot be directly constrained by the agents.

\begin{definition}[Minimally Implying Subset \cite{smith}]
A minimally implying subset $M$ of a set of predicates $S$ is any subset of $S$ such that removing any element from $M$ makes it no longer contain or imply all predicates in $S$.
\label{def:min-subset}
\end{definition}

Finally, we define compatibility, which determines whether a set of constraints can be satisfied simultaneously:

\begin{definition}[Compatibility \cite{smith}]
A set of constraints $S$ is compatible iff no constraint in $S$ is implied by more than one unique minimally implying subset of $S$.
\label{def:compat}
\end{definition}

\subsection{Problem Definition}
We may now provide a general formulation of our MT-MR-IA task allocation problem, referred to as ``Task Allocation for Multitasking robots under Physical Constraints'' 
or ``TAMPiC'' for short.

\begin{definition}[TAMPiC] An instance of TAMPiC
is a tuple $(F, Q, R, C, T, I, \Delta)$ defined as follows: 
\begin{itemize}
    \itemf The set of all unique instantiated predicates. 
    %F is a domain constant.
    \itemq A set of CIRs $\{q_l\}$. By definition, each $q_l$ takes the form $S_l \rightarrow f_l (S_l \in 2^F, f_l \in F)$. 
    \itemr A set of robots $\{r_i\}$. Each $r_i$ has a set of capabilities $\{c_j\}$ and each $c_j$ has an associated set of predicates $P_j = \{f_k \in F\}$, denoted as $c_j \rightarrow P_j$, that will become constrained if $c_j$ is activated and a cost $k_j$ expressed as a nonnegative integer which will be subtracted from solution utility if $c_j$ is activated. Each $k_j$ is assumed zero if not specified.
    \itemt A set of tasks $\{t_m\}$. Each {$t_m = (Y_m,u_m)$ where $Y_m$} is a nonempty set of predicates to be constrained and/or capabilities to be activated in order to fulfill the task, and $u_m$ is the utility for {fulfilling} the task.
    % We define $X * T$ such that it returns the set of all $t_m$ where $x_m = 1$. 
    \itemi A set of fully instantiated predicates $I \in 2^F$ representing constraints present in the initial state.
    \itemd 
    A function that returns a set of candidate initial states that are conducive to instantaneous assignment, given the problem initial state: $\Delta(I) = \{\delta\}, \delta \in 2^F$.
    %$\{\delta_1, \delta_2,...\}$; $\delta = I \mapsto I' \in 2^F$. $\Delta$ is a set of domain-dependent multitasking reconfiguration functions.
\end{itemize}
Then, the goal of the problem is to
%\begin{center}
    find $\delta_{max} \in \Delta(I)$ that maximizes the following objective:
%\end{center}
\vskip2pt
\begin{center}
$\text{maximize} \displaystyle\sum\limits_{m} x_mu_m -\displaystyle\sum\limits_{i, j} y_{i, j}k_{i, j}$\\
\end{center}
s.t.  the set of all constraints from the activated capabilities based on $\delta_{max}$ must be compatible, 
% \begin{center}
% $\text{maximize} \displaystyle\sum\limits_{h=1}^{m} x_hu_h -\displaystyle\sum\limits_{h=1}^{j} y_hk_h$\\
% \end{center}
where $x_m$ = 1 if $t_m$ is assigned or 0 otherwise and $y_{i, j}$ = 1 if $c_j$ is activated on $r_i$ or 0 otherwise. 
%s.t. the set of all constraints from activated capabilities and the chosen $\delta(I)$ must be compatible.
%We will hereafter use $\delta$* to denote a chosen $\delta$. 
%chosen to modify $I$, regardless of whether it is $\delta_{max}$ or not.

\end{definition}

The definition above bears some similarities to the planning problem. However, states can change over planning steps; for task allocation with instantaneous assignments, we only take a ``snapshot'' at a specific planning step. 
Each $\delta$ can be considered as moving the step at which the snapshot is taken. 
%{\color{orange}Furthermore, a state in TAMPiC includes only the set of predicates corresponding to the physical constraints present, instead of all predicates for specifying the world state in planning, since TAMPiC only deals with constraints in a specific snapshot.}
Note how such a definition prepares us to consider extended assignments in future work.
In addition, CIRs may appear to be similar to actions/operators in planning. From this perspective, in contrast to actions, CIRs are passively activated whenever the required preconditions are satisfied. This observation makes CIRs more akin to derived predicates or events in PDDL+~\cite{fox}, though adapted for multiple agents.

Since we consider instantaneous assignments in this work, we consider the constraints associated with a task for the entire duration.
The domain-dependent function $\Delta$ 
%represents the pre-processing step discussed earlier,
%which
is assumed to be given and applied once before task allocation to update the initial state to make it conducive to instantaneous assignment. 
%multitasking.
When multiple candidate initial states (after update) are possible, we will check for each separately. 
%If a user would like to compare multiple $\delta$*s, they may put each into $\Delta$.
%In practical applications, this may require the introduction of complementary tasks to accomplish the required changes.
Addressing time extended assignment in future work will eliminate the need for $\Delta$ entirely. 
%In the motivating example, if the boxes are both initially on the ground, applying $\delta$* would result in a new initial state where the boxes are stacked. 
Note that $\Delta$ only updates the initial state for task allocation purposes; it does not handle the steps required to stack the boxes in the physical environment.
To simplify the following discussion, we assume that the boxes are initially stacked as in Fig. \ref{fig:motivating} and applying $\Delta$ returns a set of a single element that is the problem initial state (i.e., $\Delta(I) = \{\delta\}$).
We may now give a formal specification of our running example:

\begin{center}
\vskip-3pt
    $R = \{r_1\}$, $r_1 = \{C_{Push}(r_1,X),C_{StrongPush}(r_1,X)\}$
\end{center} 
\vskip-3pt
This specifies that the single robot in the scenario has two capabilities: $Push$ and $StrongPush$, each taking the robot itself and the object to be pushed as arguments. 
We specify capabilities with their owner robot as the first argument to differentiate the same capabilities on different robots.
There are two capabilities
$C = \{C_{Push}(X,Y), C_{StrongPush}(X,Y)\}$: \\
$C_{Push}(X,Y) \rightarrow \\
\indent F_{Pos}(X) \land F_{Pos}(Y) \land \neg F_{Weight+}(Y) \land \neg F_{On}(Y,Z)$ \\
$C_{StrongPush}(X,Y) \rightarrow F_{Pos}(X) \land F_{Pos}(Y) \land \neg F_{On}(Y,Z)$

The positions of both the pusher robot and the object being pushed are constrained, and the object being pushed must be not on top of another object.
$Push$ also requires the object to be not heavy (i.e., $\neg F_{Weight+}(Y)$). 
$T = \{t_1, t_2\}$:
{
\begin{center}  %\end{center} 
$t_1 = (\{F_{Pos}(o_1)\}, u_1 = 1)$, %\end{center}
$t_2 = (\{F_{Pos}(o_2)\}, u_2 = 3)$\end{center}
} %put these on two lines instead of three
There are two tasks considered. The first task requires the position of a specific object ($o_1$) to be constrained (i.e., it must be moved to a specific position). The utility is 1.
The second task is similarly defined with utility 3. There are two CIRs $Q = \{q_1,q_2\}$: \\
{%\begin{center}
$q_1 = \{F_{On}(X,Y), F_{Pos}(Y)\} \rightarrow F_{Pos}(X)$ \\ %\end{center} \\
$q_2 = \\
\text{\,\,\,} \{F_{On}(X,Y), F_{Weight}(Y), F_{Weight}(X)\} \rightarrow F_{Weight+}(Y)$
%\end{center}
} %Put these on three lines instead of four

The first rule specifies that $X$ being on $Y$ constrains $X$'s position if $Y$'s position is constrained. 
This rule is used to infer that a stack of objects can be pushed by pushing only the bottom object, resulting in task synergies. See Sec. \ref{sec:running} Item 1).
The second rule specifies that if $X$ is on top of $Y$, and both objects have a weight, the effective weight of $Y$ is increased.
This rule suggests the additional constraints introduced by multitasking. See Sec. \ref{sec:running} Item 2). The initial state is:
% Define initial state:

\begin{center}
\vskip-3pt
$I = \{F_{On}(o_2,o_1),F_{Weight}(o_1),F_{Weight}(o_2)\}$\end{center} 
\vskip-3pt
In the initial state, $o_2$ is on top of $o_1$; both objects have a weight.
%Note that 
This means the robot cannot directly push $o_2$ because it is out of reach.
Single-tasking approaches would fail to recognize that $t_2$ can be completed at all.
%However, 
In TAMPiC,
this can be accomplished by activating the $C_{StrongPush}(r_1,o_1)$ {capability to push both boxes simultaneously}.% to achieve multitasking after considering its synergistic and restrictive aspects. 

\subsection{Solution Method}
\label{sec:solution}
%The MT-MR-IA problem 
The MT-MR-IA problem without considering physical constraints, 
equivalent to the well-known NP-complete Set Covering Problem~\cite{gerkey},
is a special case of TAMPiC when $Q = \emptyset$ and $\Delta(I) = \{I\}$. %* is the identity map. 
Thus, TAMPiC is NP-hard. 
Despite the computational challenge, we explore a general process for
converting instances of TAMPiC to Weighted MAX-SAT, which is a well-studied NP-complete problem and has efficient solutions in practice.
The converted problem can be given to any Weighted MAX-SAT solver. 
We provide a definition of Weighted MAX-SAT, similar to \cite{heras}:

\begin{definition}
The Weighted MAX-SAT problem is a tuple $(V, S, W)$ where $V$ is the set of variables, $S$ is a set of disjunctive clauses of literals of $V$, and $W$ is a set of non-negative weights assigned to each clause in $S$.
\label{def:maxsat}
\end{definition}

A literal refers to a variable $v$ or its negation $\neg v$.
The goal of Weighted MAX-SAT is to find truth values for variables in $V$ to maximize the combined weight of the satisfied clauses in $S$. 
%We use the notation 
{\color{black}$(s, w)$ denotes a clause and its associated weight.}

% \begin{definition}[Rule Density]
%     The rule density $\rho$ of an instance of TAMPiC is the smallest value such that no CIR in the problem can be instantiated in more than $\rho$ unique ways.
%     \qed
%     \label{def:info-den} 
% \end{definition}

The full process of conversion and querying the SAT solver for a solution to a TAMPiC problem is referred to as SAT-based Task Assignment with Multitasking Robots, or STAMR.
Throughout the conversion, unless otherwise specified, every clause has weight $\alpha$ that is strictly greater than the sum of the utilities of all tasks.
$\alpha$ is used to ensure the satisfaction of these clauses.
Alternatively, many Weighted MAX-SAT solvers allow for ``hard'' clauses, which must be satisfied.
Either method may be used in practice.
{$F, C, I,$ and $\Delta$}: $F$ and $C$ are redundant  given the other components for ease of reference so we do not discuss their conversion. 
%and used for ease of reference, and %come from the domain, 
For I, and $\Delta$, We will need to apply the conversion for %$\delta$*$(I)$ is applied. 
%To choose $\delta$*, we solve the problem once for 
each candidate initial state in $\Delta(I)$.
%Thus, no conversion is needed for $F$, $C$, $I$, or $\Delta$.

%\textbf{I$^{\prime}$:} Each predicate in $I' = \delta$*$(I)$ introduces its own clause.
{$\delta$:} Each predicate in $\delta, \delta \in \Delta(I),$ introduces its own clause.

$Q$:
To convert CIRs, a clause for each possible instantiation of each CIR must be created.
We will use a certain Boolean identity extensively for the entire STAMR conversion process: $(p \rightarrow q) \equiv (\neg p \lor q)$.
We use this equivalence to ensure that the Weighted MAX-SAT formula is in the conjunctive normal form (CNF), which is required for many Weighted MAX-SAT solvers.
For example, one instantiation of $q_1$ 
% \begin{center}
% $\{F_{On}(X,Y), F_{Pos}(Y)\} \rightarrow F_{Pos}(X)$
% \end{center}
is:

\begin{center}
\vskip-3pt
$\{F_{On}(o_1,o_2), F_{Pos}(o_2)\} \rightarrow F_{Pos}(o_1)$ 
\end{center}
\vskip-3pt
and the corresponding weighted clause would be:

\begin{center}
\vskip-3pt
% (($\neg (F_{On}(o_1,o_2) \land F_{Pos}(o_2)) \lor F_{Pos}(o_1)) \equiv 
$(\neg F_{On}(o_1,o_2) \lor \neg F_{Pos}(o_2) \lor F_{Pos}(o_1))), \alpha)$
\end{center}
\vskip-3pt
% \TD{Should I be using the $\equiv$ symbol here? May be interpreted as creating a tautological clause, rather than me saying "this is equivalent to."}

$R$:
Each $r_i$ is defined as a set of capabilities.
To convert capabilities, create a clause for each instantiated capability that requires the predicates constrained by that capability to be true if the capability is activated. 
For example, robot $r_1$ has a capability 
%\begin{center}
$C_{StrongPush}(r_1,Y)$,  
%\end{center}
%which constrains (among other things) the position of $r_1$, a
and one instantiation of this capability might be
%\begin{center}
$C_{StrongPush}(r_1,o_1)$.
%\end{center}
%and if the robot's position is the only predicate constrained by activating the capability, 
%Then the corresponding Weighted MAX-SAT clause would be:
Then the instantiated capability can be converted to:
\begin{center}
\vskip-3pt
$((\neg C_{StrongPush}(r_1,o_1) \lor (F_{Pos}(r_1) \land F_{Pos}(o_1) \land \neg F_{On}(o_1,o_2)), \alpha)$
\end{center}
\vskip-3pt
which can then be converted to multiple disjunctive clauses using the distributive law, each with weight $\alpha$.

To handle capability costs, however, we must make use of the following theoretical result since MAX-SAT does not natively support costs (negative weights).
First, we introduce the following notations.
Let $\beta = \{\{s_1,w_1\},\{s_2,w_2\},...\}$ for a given weighted MAX-SAT problem.
Note that $V$ is implied given $S$ and $W$ in Def. \ref{def:maxsat}.
We define $\rho(\{s_i,w_i\}) = \{\neg s_i, -w_i\}$
%and use $L$ to  denote the list of all predicates in $\beta$. 
and denote an interpretation of all the predicate variables in $\beta$  as $X$. 
We further define $opt(\beta,X) = 1$ if $X$ is an optimal solution to $\beta$ or 0 otherwise.
Let 
$\beta'$ be $\beta$ after applying $\rho$ to an element in $\beta$.

% Given:

% \begin{center}

% $\beta : \{\{s_1,r_1\},\{s_2,r_2\},...\}$,

% $s_i$: Each $s_i$ is any valid SAT clause,

% $r_i$: The reward for satisfying $s_i$,

% $\rho$: $\rho(s_i,r_i) = \{\neg s_i, -r_i\}$,

% $L:$ A list of all predicates present in all $s_i \in \beta$ denoted $\{\ell_1, \ell_2,...\}$,

% $X:\{x_1, x_2,...\}:x_i \in L \forall i$ (X is any interpretation of $\beta$),

% $opt(\beta,X) = 1$ if X is an optimal solution to $\beta$ or 0 otherwise, and

% $\beta': \beta$ after applying $\rho$ to any one $s_i \in \beta$,

% \end{center}

\begin{theorem}
    $opt(\beta,X)=opt(\beta',X)$.
\end{theorem}

\begin{proof}
    Let $V(\beta,X) = \displaystyle\sum w_i: X \models s_i$, and let $s_j$ be the clause that satisfies $s_j \in \beta \land s_j \notin \beta'$. %($s_j$ is the transformed clause before $\rho$ is applied). \\
    There are two possible cases here. 
    %\begin{center}
    %$\begin{math}
    1) Case 1: $X \models s_j \rightarrow V(\beta', X) = V(\beta, X) - w_j$; 2) Case 2: $X \cancel \models s_j \rightarrow V(\beta', X) = V(\beta, X) + (-w_j)$. Hence, in both cases we have 
    $V(\beta', X) = V(\beta, X) - w_j$. Since $w_j$ is a constant across all $X$ under any given  MAX-SAT problem with a chosen $s_j$, we have 
    $opt(\beta, X) = opt (\beta', X)$.  %\forall \beta,X
    % \end{math}
    % \end{center}
    % {\color{blue} You had a comment here saying to replace the $\therefore$ marks with things like "thus," or "so we have," but that's the literal meaning of the mark.}
\end{proof}

% \begin{proof}
%     Let $\rho(\beta,X) = \displaystyle\sum r_i\in \beta : X \models s_i$, let $s_j : s_j \in \beta, s_j \notin \beta'$ ($s_j$ is the transformed clause before $\rho$ is applied). \\
%     There are then two cases:
%     \begin{center}
%     \begin{math}
%     (1) (X \models s_j) \rightarrow (\rho(\beta', X) = \rho(\beta, X) - r_j) \\
%     (2) (X \cancel \models s_j) \rightarrow (\rho(\beta', X) = \rho(\beta, X) + (-r_j) \\
%     \therefore \rho(\beta', X) = \rho(\beta, X) - r_j\\
%     \text{Note: $r_j$ is a constant for any given F across all X.}\\
%     \therefore opt(\beta, X) = opt (\beta', X) \forall \beta,X
%     \end{math}
%     \end{center}
%     {\color{blue} You had a comment here saying to replace the $\therefore$ marks with things like "thus," or "so we have," but that's the literal meaning of the mark.}
% \end{proof}

We can apply the above theorem multiple times to apply $\rho$ to multiple clauses.
Note that negating a clause in CNF causes that clause to potentially no longer be in CNF.
While it is trivial to convert the negated clause back to CNF by breaking it into multiple clauses, the weight of each resulting clause cannot be easily resolved.
To address this problem, we attach the weight to a marker clause for each instantiation of each capability to denote if it is activated:

\begin{center}
\vskip-3pt
    $\{c_{i, j}, -k_{i, j}\}$
\end{center}
\vskip-3pt

% \begin{center}
%     $\{c_{j,l}, -k_j\}$
% \end{center}

The marker clauses are then transformed according to $\rho$.
%We use the second subscript to index into instantiations.
Note the change of sign above with respect to the original problem.

% \change{The weight of each clause is said to be $\alpha$ by default. OK to remove the specification here?}

% To comply with the requirement for our formula to be in CNF, we create a clause for each predicate constrained by each capability, rather than a single clause for each capability which covers all predicates constrained by that capability.
% For example, in addition to the clause outlined above, the following three clauses would be created when converting our running example, representing the other constraints created by activating $C_{StrongPush}(r_1,o_1)$:

% \begin{center}
% $((\neg C_{StrongPush}(r_1,o_1) \lor F_{Pos}(r_1)), \alpha),$

% $((\neg C_{StrongPush}(r_1,o_1) \lor F_{Pos}(o_1)), \alpha),$

% % $((\neg C_{StrongPush}(r_1,o_1) \lor \neg F_{On}(o_1,o_2)), \alpha),$

% $((\neg C_{StrongPush}(r_1,o_1) \lor \neg F_{On}(o_1,r_1)), \alpha)$.
% \end{center}

$T$:
To convert tasks, there are two steps for each task. 
First, create a clause with weight equal to the task's utility and which contains only an assignment literal corresponding to whether the task is assigned. 
For example, for $t_1$, the corresponding Weighted MAX-SAT clause would be:

\begin{center}
\vskip-3pt
$((t_1), 1)$
\end{center}
\vskip-3pt

Then, similar to $Q$, create a clause for each instantiation of each task, which forces the predicates for that instantiation to be true, and also another clause which forces at least one of the instantiations to be satisfied if the task's assignment literal is true. In the case of $t_1$, it only has a single instantiation.

\begin{center}
\vskip-3pt
    $((\neg t_{1,1} \lor F_{Pos}(o_1)), \alpha),$
    $((\neg t_1 \lor t_{1,1}), \alpha)$
\end{center}
\vskip-3pt

\textbf{Finalizing}: 
To prevent constraints being created without the corresponding capabilities or CIRs being activated to create them, each constraint must imply at least one of the ways in which it could have been generated.
For example, we have $((\neg F_{Pos}(o_2) \rightarrow (C_{Push}(r_1, o_2) \lor C_{StrongPush}(r_1, o_2) \lor q_1)), \alpha)$, which can be converted into:
\begin{center}
\vskip-3pt
    $((\neg F_{Pos}(o_2) \lor C_{Push}(r_1,o_2) \lor (C_{StrongPush}(r_1,o_2) \lor (F_{On}(o_1,o_2) \land F_{Pos}(o_2)), \alpha)$
\end{center}
\vskip-3pt
% Note that this is essentially equivalent to $((\neg F_{Pos}(o_2) \rightarrow (C_{Push}(r_1, o_2) \lor C_{StrongPush}(r_1, o_2) \lor q_1)), 

Additionally, each CIR and capability must set all other CIRs and capabilities {(including different instantiations)} that constrain the same predicates to false to 
%capture incompatibility:
prevent incompatibility (Def. \ref{def:compat}). For example:  

\begin{center}
\vskip-3pt
    $((\neg C_{Push}(r_1, o_2) \lor (\neg C_{StrongPush}(r_1, o_2) \land \neg (F_{On}(o_1,o_2) \land F_{Pos}(o_2))), \alpha)$
\end{center}
\vskip-3pt

The number of clauses in the Weighted MAX-SAT problem corresponding to each TAMPiC instance is on the order of the $k$-permutation of the set of possible instantiations for each argument to any predicate in $F$, where $k$ is the maximum number of arguments any predicate takes.
Because %$F$, which defines 
$k$ is a domain constant, the conversion %STAMR 
is polynomial w.r.t. the size of the TAMPiC instance.
%TAMPiC remains NP-hard, however, because Weighted MAX-SAT is NP-complete.
The converted problem can be given to any Weighted MAX-SAT solver and the resultant solution gives us a solution to the original TAMPiC problem.
% Any $C_i$ literal set to true, along with its instantiation, implies that the corresponding capability should be instantiated and activated in TAMPiC and vice versa. %the same way.
%Thus, we may convert any instance of TAMPiC to Weighted MAX-SAT and solve it.

\begin{theorem}
    The compilation solution is both sound and complete for TAMPiC, assuming the Weighted MAX-SAT solver used is also both sound and complete.
\end{theorem}

This is correct by construction.
Each element of TAMPiC is converted to Weighted MAX-SAT, and no element is left out.
Predicates can only be constrained by the activation of capabilities or CIRs.
Because activating a constraint or CIR prevents the activation of any other capabilities or CIRs constraining the same predicate, compatibility is preserved.

The solution to the generated Weighted MAX-SAT problem implies a solution {to the %original
TAMPiC problem.}
For each literal corresponding to an instantiated capability, if that literal is true, then activating the corresponding capability with that instantiation is part of the {solution to the %original
TAMPiC problem.}
%Conversely, g
Given a solution to the TAMPiC problem, setting the literals corresponding to the instantiations of activated capabilities to true will solve the converted Weighted MAX-SAT problem.
This will also require activating any relevant CIRs and setting task literals to true if the corresponding tasks were assigned.
Literals corresponding to instantiations of CIRs and capabilities not activated should be set to false.

\begin{corollary}
    TAMPiC is NP-complete.
\end{corollary}

It follows immediately given that TAMPiC is in NP and a solution to TAMPiC is a solution to Weighted MAX-SAT and vice-versa based on the conversion (reduction). %compilation/reduction.

\subsection{Greedy Heuristic}
We also propose an approximation method using a simple greedy heuristic. 
It chooses the task with the highest utility, and only converts TAMPiC with that single task into the corresponding weighted MAX-SAT problem and solves it.
Then, it retrieves any implied constraints from the solution, and incorporates these constraints into future iterations as clauses with weight $\alpha$, which simply state the presence of these constraints.
This process is repeated 
%until the SAT solver has given a SAT formula 
for each task, ordered from the highest utility to the lowest.
Any tasks that were successfully fulfilled are fulfilled in the same way in the returned solution for the original problem.
This method is referred to as STAMR-Greedy, or STAMR-G for short.
While we discuss no theoretical 
%guarantees
properties of STAMR-G's, either on its running time or solution-optimality, 
%practical 
empirical results show that it outperforms the baseline.

% Note that none of these guarantees apply to the greedy approximation method outlined in the next section.
% The only guarantee provided for that method is that if at least one task is assignable, at least one task will be assigned.
% However, in practice, the greedy method performs reasonably well. 

\section{Results}
\vskip-3pt
\begin{figure*}[t!]
    \vskip0pt
    \centering
    \includegraphics[width=0.6\columnwidth]{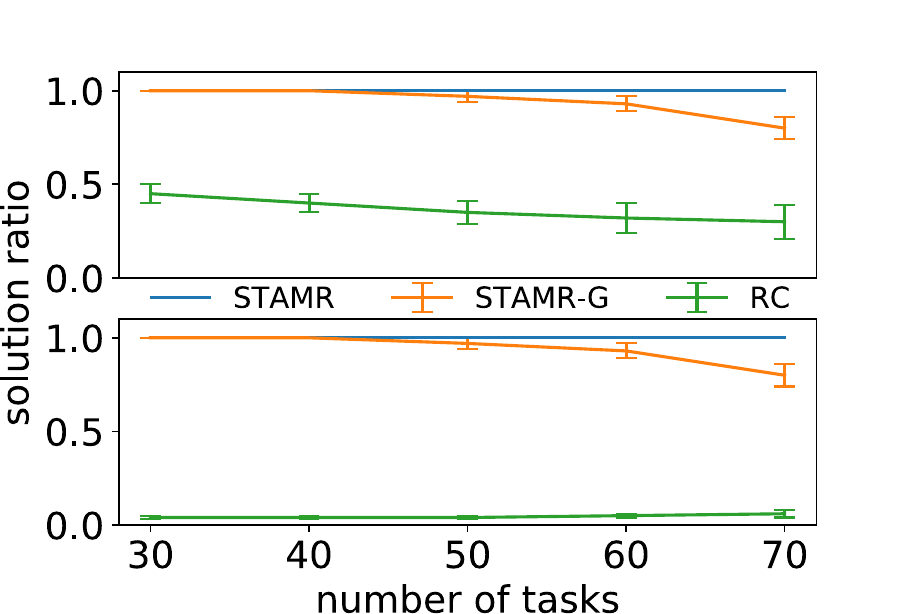}
    \hfill
   \includegraphics[width=0.6\columnwidth]{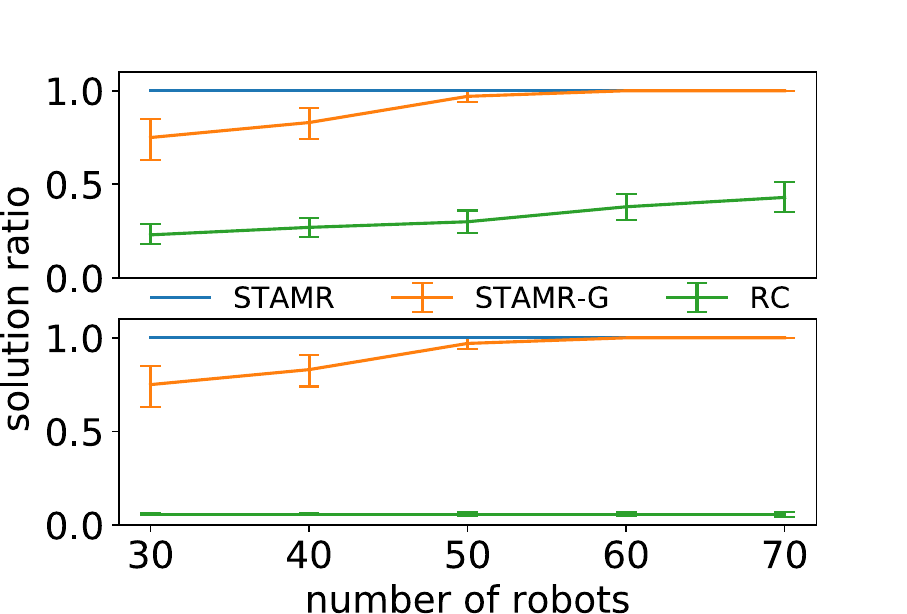}
    \hfill
    \includegraphics[width=0.6\columnwidth]{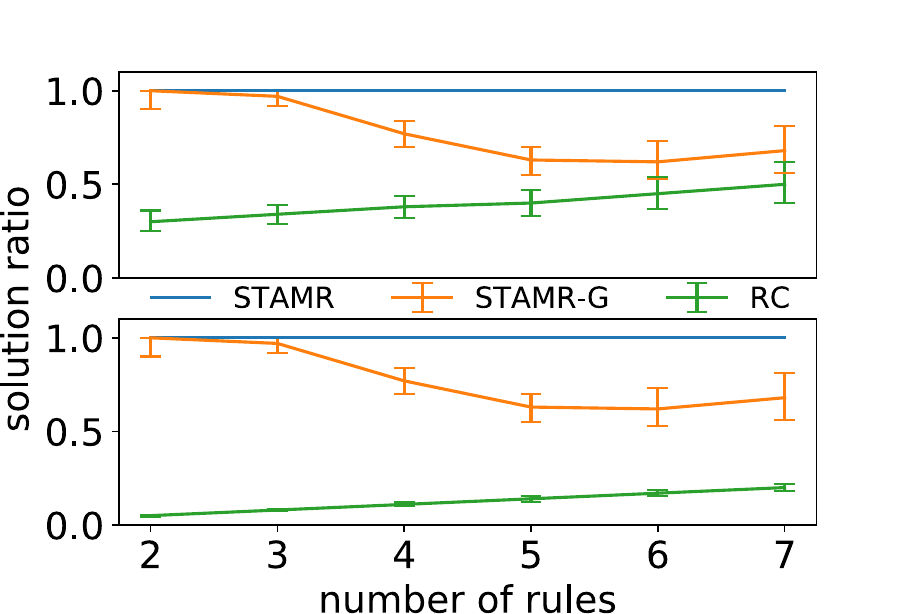}
    \hfill
    \vskip-5pt
    \caption{Solution ratio as \# of tasks (left),  robots (center), and  CIRs (right) increases in first (top) and second (bottom) setting.}
    
    \vskip-13pt
    \label{fig:accuracy}
\end{figure*}

\begin{figure*}
    %\vskip15pt
    \includegraphics[width=0.6\columnwidth]{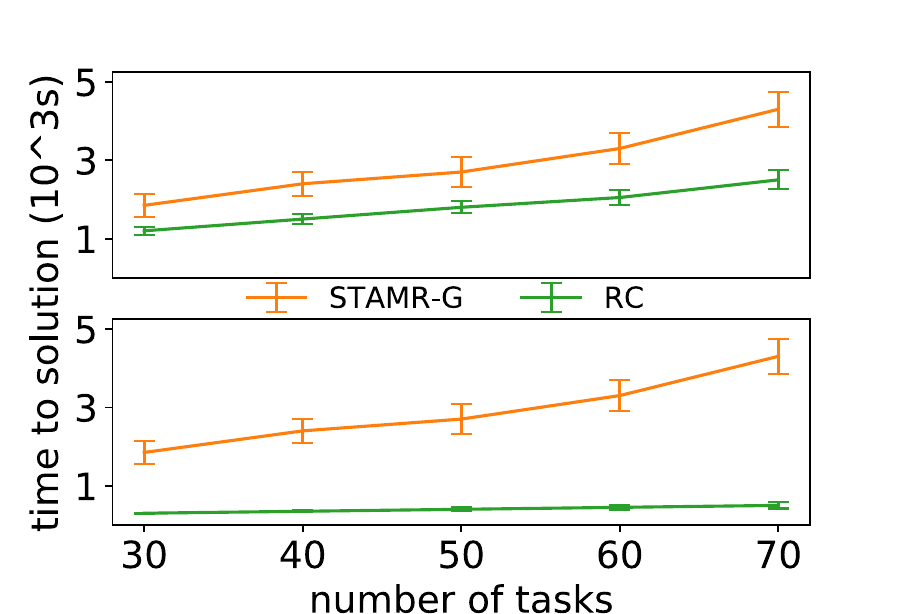}
    \hfill
    \includegraphics[width=0.6\columnwidth]{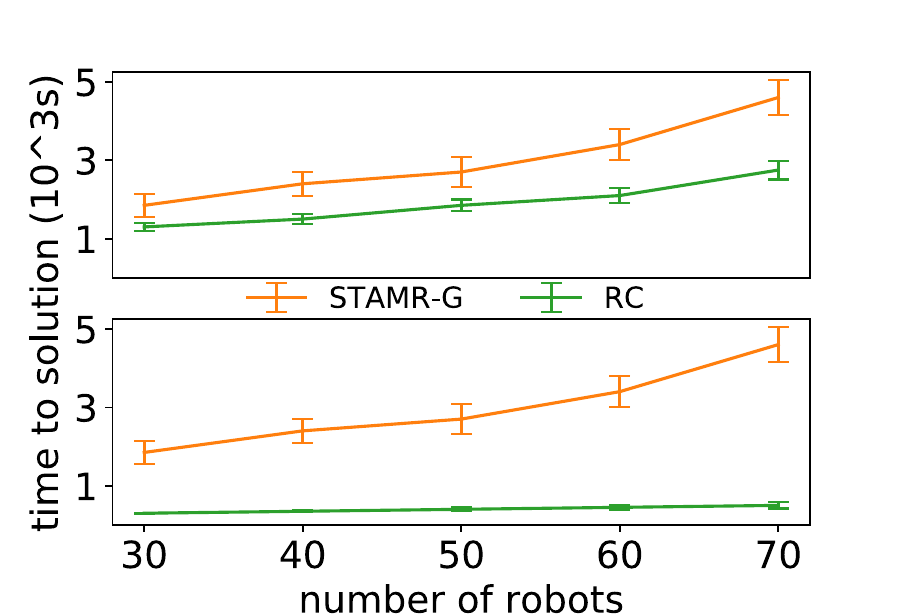}
    \hfill
    \includegraphics[width=0.6\columnwidth]{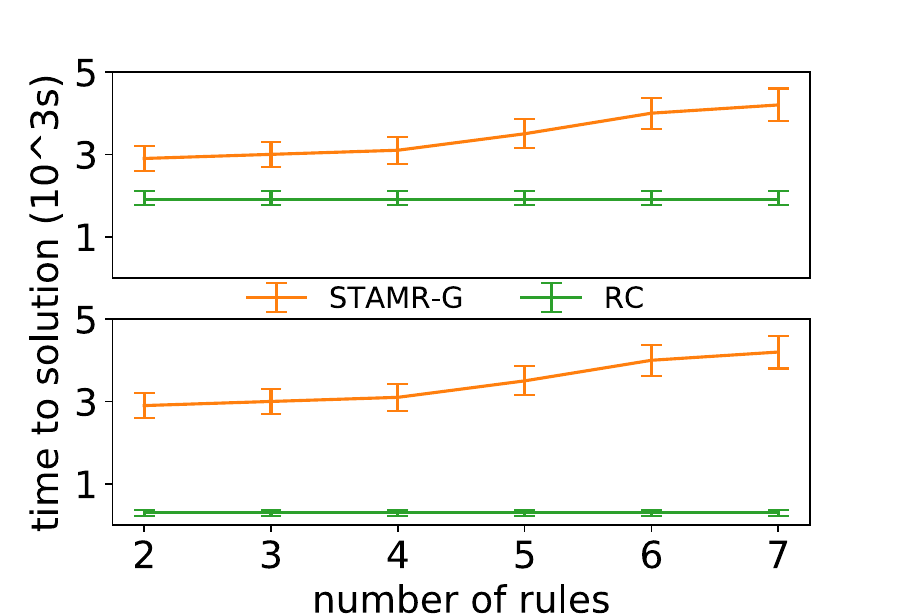}
    \hfill
    \vskip-5pt
    \caption{Time taken as \# of  tasks (left),  robots (center), and  CIRs (right) increases in the first (top) and second (bottom) setting. The solution time for STAMR is not plotted since it required substantially more time than the others.
    }
    
    \vskip-10pt
    \label{fig:time}
\end{figure*}

\begin{figure*}[t!]
    \vskip0pt
    \centering
    \includegraphics[width=0.45\columnwidth]{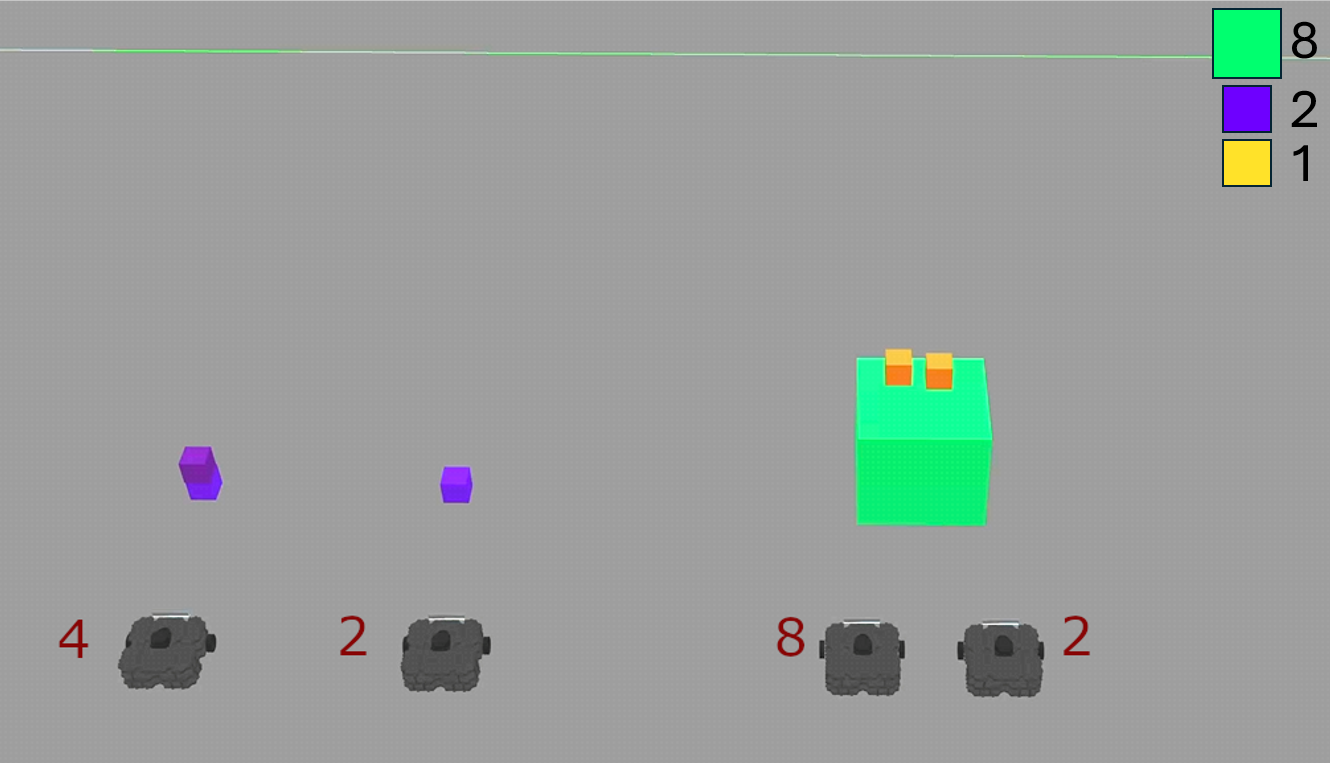}
    \hfill
    \includegraphics[width=0.45\columnwidth]{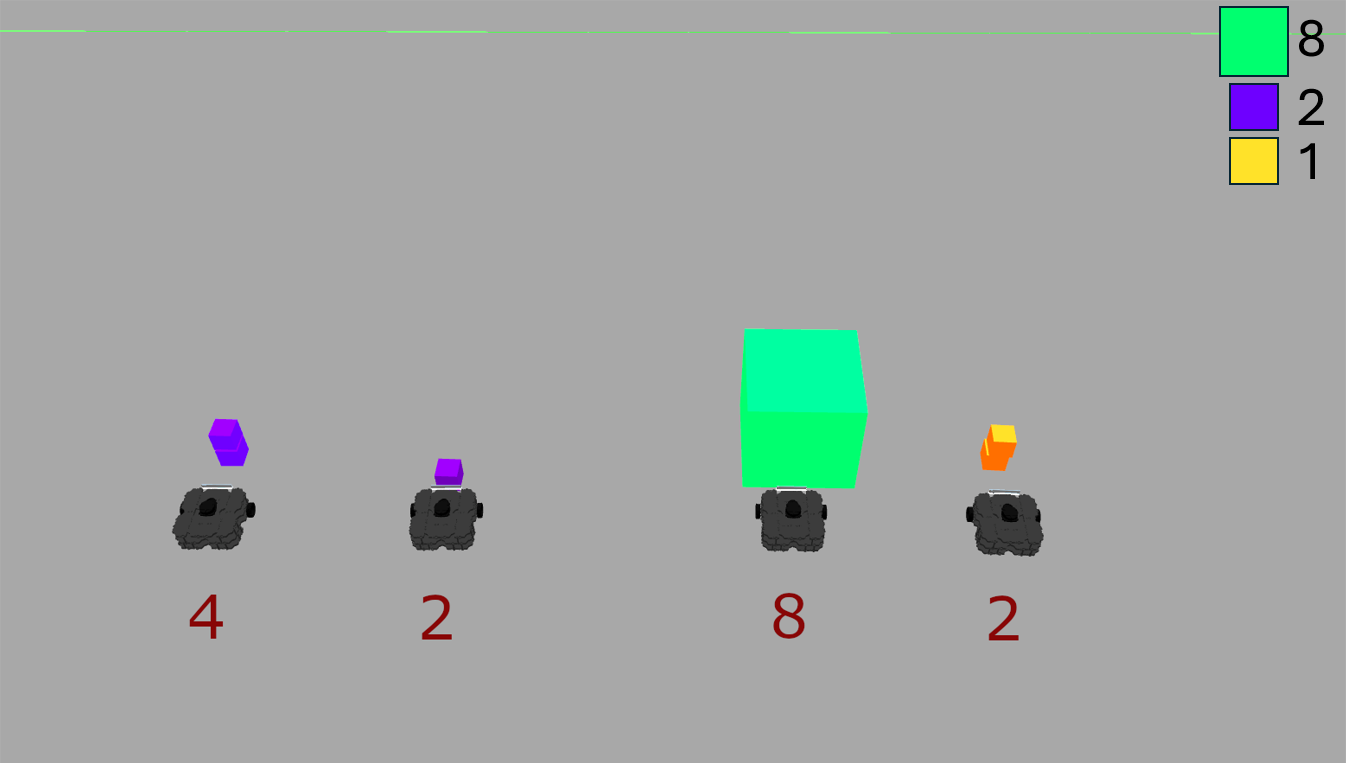}
    \hfill
    \includegraphics[width=0.45\columnwidth]{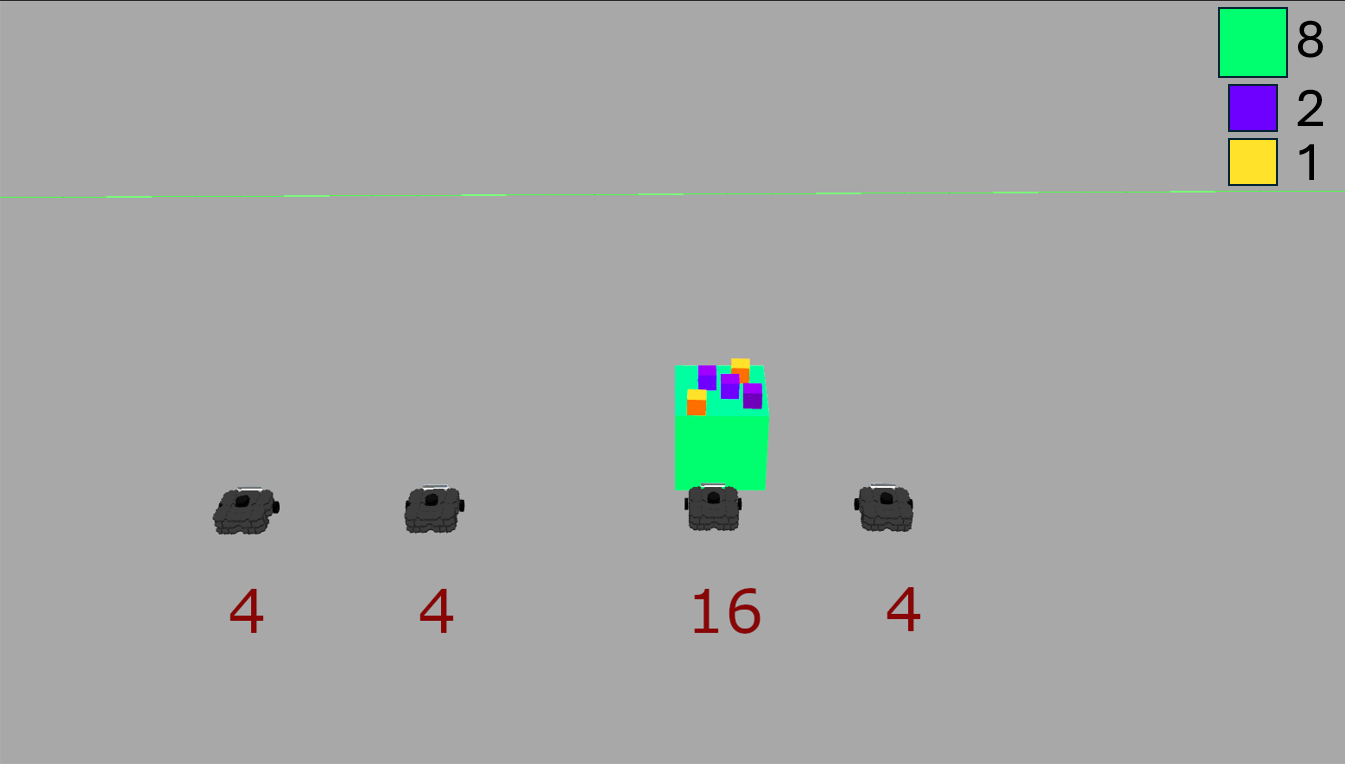}
    \hfill
    \includegraphics[width=0.45\columnwidth]{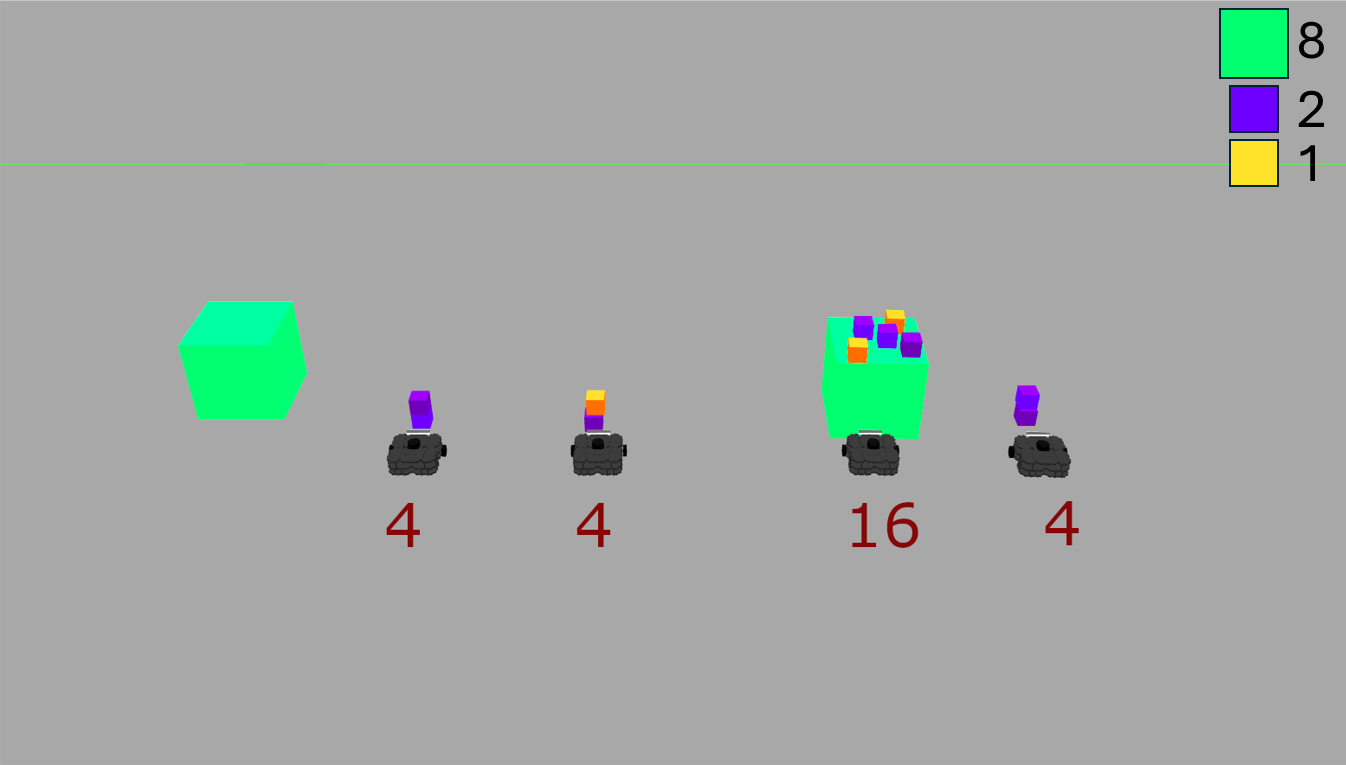}
    \vskip-5pt
    \caption{
    Four site clearing scenarios with one of their optimal solutions illustrated, respectively.  
     % The simulation environment with the first set of initial conditions, when the robots have just lined up to perform their tasks (leftmost). The same time, for the second initial conditions (second). The same time, for the third initial conditions (third). The fourth time, with the final set of initial conditions (rightmost).
     }
     \vskip-10pt
    \label{fig:experiment}
\end{figure*}

\begin{figure*}[t!]
    \vskip0pt
    \centering
    \includegraphics[width=0.6\columnwidth, height=0.3\columnwidth]{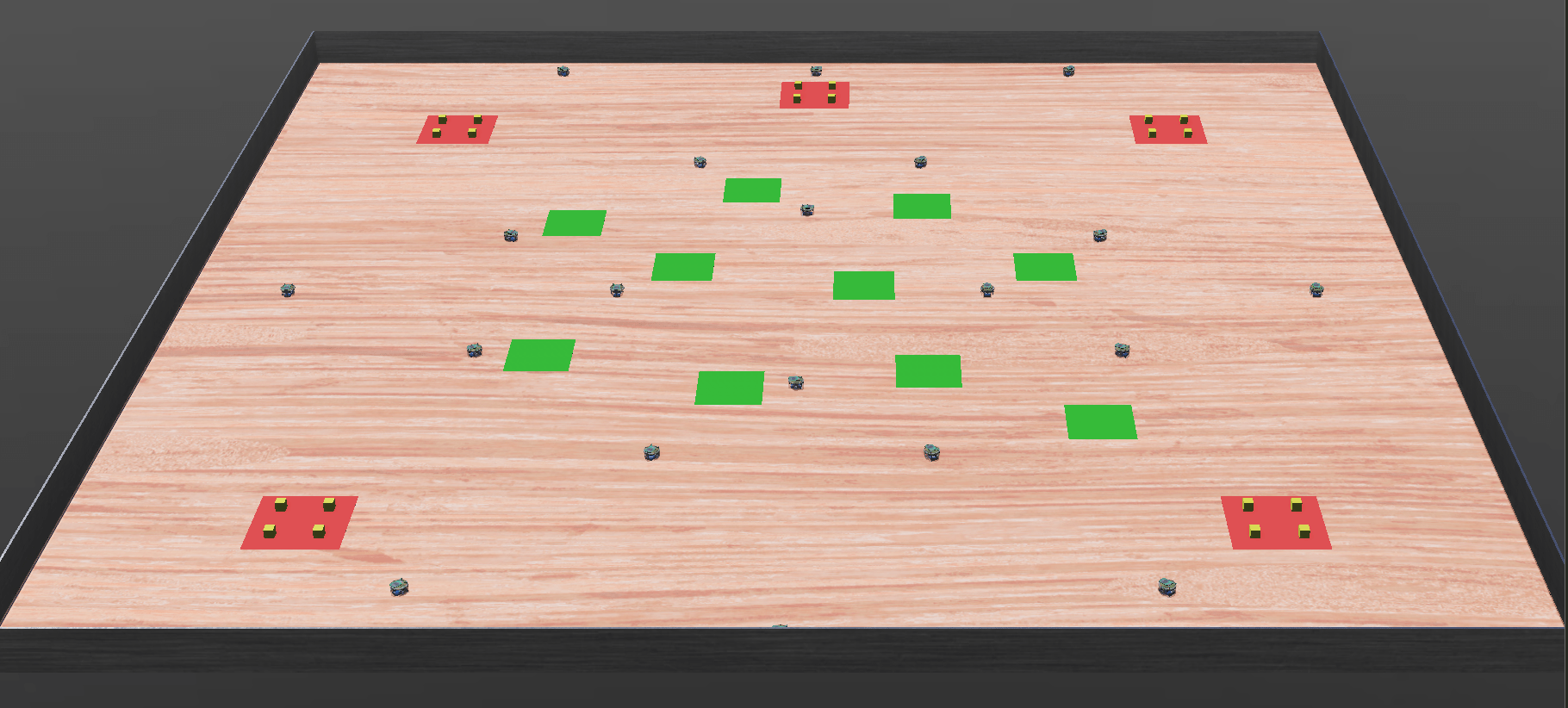}
    \hfill
    \includegraphics[width=0.6\columnwidth, height=0.3\columnwidth]{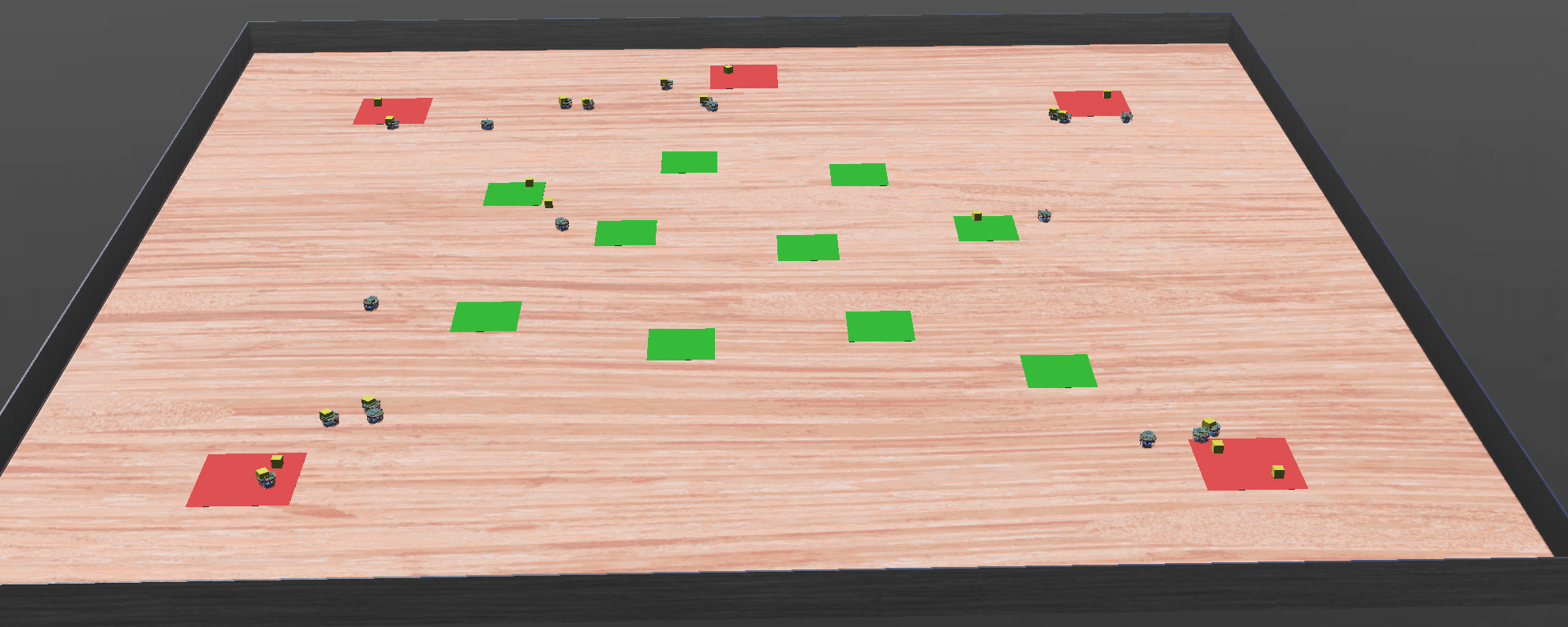}
    \hfill
    \includegraphics[width=0.6\columnwidth, height=0.3\columnwidth]{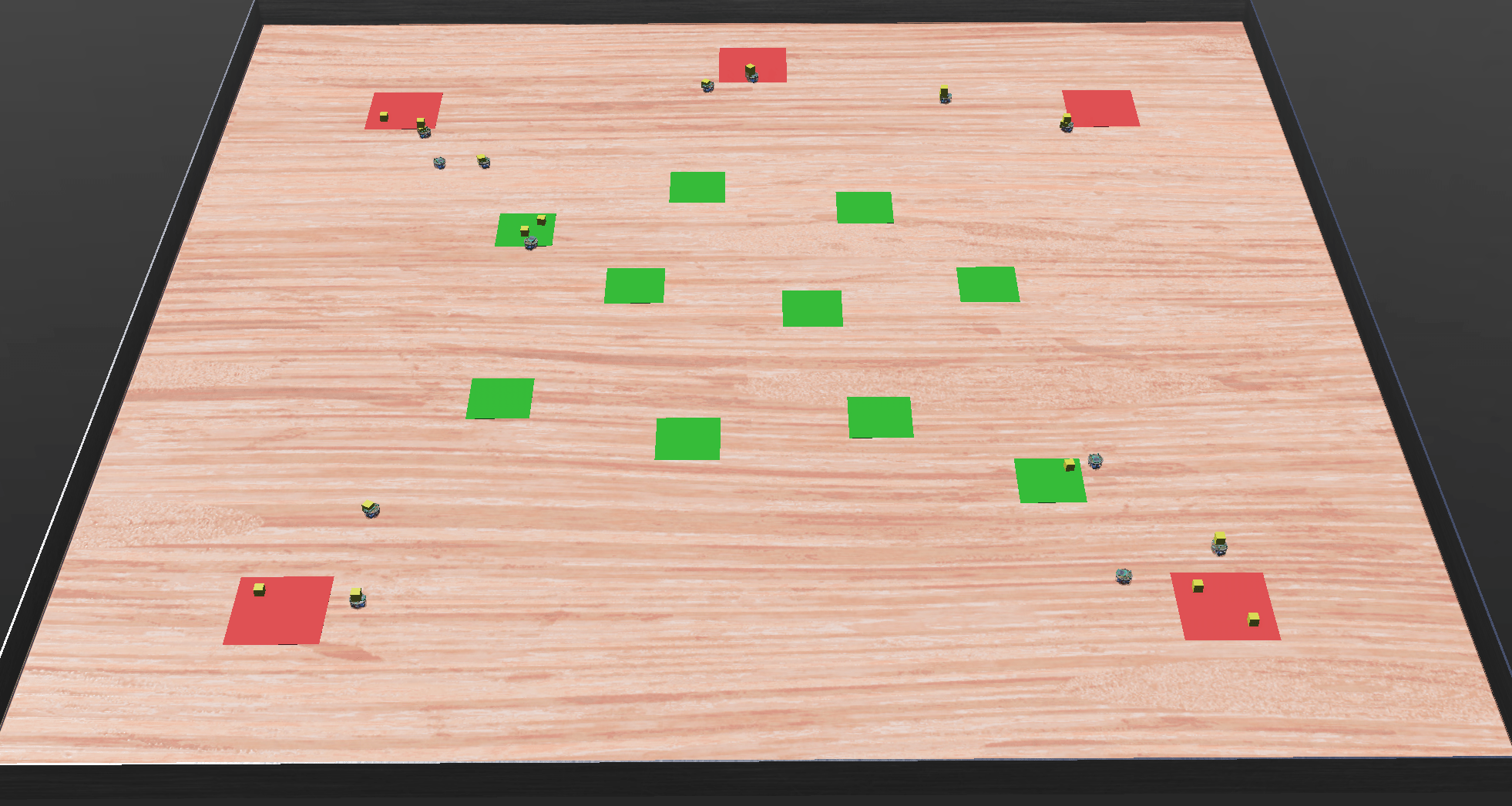}
    \hfill
    \caption{
     The product delivery scenario in its initial state. The entire simulation environment (left). The baseline, midway through execution; note the clumps of robots (center). Our congestion-aware approach (right).}
    \vskip-20pt
    \label{fig:experiment2}
\end{figure*}

For evaluation, we first evaluate our methods in synthetic domains and compare with a baseline to validate the benefits of multitasking. 
We then further illustrate the complex task interaction among the robots using a site-clearing scenario in simulation, similar to our running example. 
Finally, we demonstrate a higher-complexity simulation to show the scalability and applicability of our approach.
In all cases, we only list key components of the problem for brevity.
% {\color{red} Need update once everything following is updated.}
% {\color{purple} Does it need updated? Seems okay to me.}

\subsection{Synthetic Evaluation}

% \change{I can't get the text coloring to work properly. The problem definition below is all new. This is what part of a randomly-generated problem would look like.}

% $R = \{r_1\}$, $r_1 = \{C_{1}(X,Y),C_{2}(X,Y)\}$, 

% $C = \{C_{1}(X,Y), C_{2}(X,Y)\} $,

% $C_{1}(X,Y) \rightarrow F_{3}(X) \land F_{3}(Y) \land \neg F_{4}(Y) \land \neg F_{6}(Y,Z)$, 

% $C_{2}(X,Y) \rightarrow F_{3}(X) \land F_{3}(Y) \land \neg F_{6}(Y,Z)$, 

% $T = \{t_1, t_2\}$, %\end{center} 

% $t_1 = (\{F_{3}(o_1)\}, u_1 = 1)$, %\end{center}

% $t_2 = (\{F_{3}(o_2)\}, u_2 = 3)$,

% $Q = \{q_1,q_2\}$, %\end{center} 

% $q_1 = \{F_{6}(X,Y), F_{3}(Y)\} \rightarrow F_{3}(X)$, %\end{center}

% $q_2 = \{F_{6}(X,Y),$ $F_{5}(Y),$ $F_{5}(X)\} \rightarrow$ $F_{4}(Y)$, 

% $I = \{F_{6}(o_2,o_1),F_{5}(o_1),F_{5}(o_2)\}$

Our baseline is the \textit{ResourceCentric} method from \cite{zhang2013considering} that assumes single-tasking robots since it has been shown to outperform other commonly used methods for multi-robot task allocation with IA. 
We will use ``RC''
%as a shorthand 
to refer to this method.
{Because our problem definition}
%is unique in a way that
poses a hurdle to running RC, 
we evaluated RC in two different settings.
%More specifically, 
The issue here is that tasks in TAMPiC may be specified as requirements for %whether or not a given predicate is constrained, 
constraints, such as moving the box expressed as a constraint on the position of the box. % in the running example. %and/or capabilities to be activated.
RC is only able to consider tasks that only require capabilities to be activated
since dealing with constraints requires CIRs. 
Hence, 
we evaluated RC in two settings. % for each evaluation. 
In the first setting, tasks generated did not require any constraints, whereas in the second setting, each requirement of each task had an equal chance to be a capability or a constraint. %changed from talking about 50% and 50%
For the second setting, because RC cannot consider tasks with constraints, all such tasks are discarded before running RC.

We analyzed randomly generated problem instances.
Because calculating the optimal solution is infeasible for large problems, we limited our examination to medium-sized problem instances. %, as defined below.
First, at most five unique uninstantiated predicates were generated ($1 \leq |F| \leq 5$), each with one or two arguments.
Each problem had $|T| = 50$, each $t_m$ with $1 \leq |Y_m| \leq 3$ and $1 \leq u_m \leq 30$.
Each problem had $|R| = 50$, each $r_i$ with $1 \leq |\{c_j\}| \leq 3$, 
%$1 \leq |C| \leq 3$, 
and each type of capability constrained one or two predicates ($1 \leq |P_j| \leq 3$).
Each problem had $|Q| = 2$, each $q_l$ with $1 \leq |S_l| \leq 3$.
$\Delta(I) = \{I\}$. %* was the identity mapping; no changes to $I$ were performed.
In all cases where a range of possible outcomes is given, random generation was used with equal chance given to each outcome.
For each evaluation setting, we used the parameter ranges as discussed above by default, {except for the parameter that we chose to vary.} % to show its impact on the solutions.
%An example of a randomly generated problem with small robot and task sizes for illustration purposes could look like the following:

% $R = \{r_1\}$, $r_1 = \{C_{1}(X,Y),C_{2}(X,Y)\}$, 

% %$C = \{C_{1}(X,Y), C_{2}(X,Y)\} $,

% $C_{1}(X,Y) \rightarrow F_{3}(X) \land F_{3}(Y) \land \neg F_{4}(Y) \land \neg F_{6}(Y,Z)$, 

% $C_{2}(X,Y) \rightarrow F_{3}(X) \land F_{3}(Y) \land \neg F_{6}(Y,Z)$, 

% %$T = \{t_1, t_2\}, %\end{center} 
% $t_1 = (\{F_{3}(o_1)\}, u_1 = 1), %\end{center}
% t_2 = (\{F_{3}(o_2)\}, u_2 = 3)$,

% %$Q = \{q_1,q_2\}$, %\end{center} 

% $q_1 = \{F_{6}(X,Y), F_{3}(Y)\} \rightarrow F_{3}(X)$, %\end{center}

% $q_2 = \{F_{6}(X,Y),$ $F_{5}(Y),$ $F_{5}(X)\} \rightarrow$ $F_{4}(Y)$, 

% $I = \{F_{6}(o_2,o_1),F_{5}(o_1),F_{5}(o_2)\}$

% %$\Delta{I} = \{I\}$

%For our synthetic evaluation, w
We tested the effects of increasing the number of tasks, robots and CIRs on the three methods (RC, STAMR, STAMR-G) in both settings.
%using the above parameters unless otherwise specified.
%For the purpose of our analysis, 
We define ``solution ratio'' %is defined
as the utility achieved by the given approach divided by the utility achieved by the optimal solution.
For each data point, $100$ randomized runs were performed. 
The results are in Fig. \ref{fig:accuracy} and Fig. \ref{fig:time}.
Note that the difference in settings had a negligible effect on the performance of STAMR and STAMR-G; this is expected since 
STAMR and STAMR-G handle capability- and constraint-based tasks {identically}. 
As more tasks were added, the relative performance of STAMR-G and RC dropped, likely due to the increased difficulty in optimization given that more tasks were available but only some of them could be assigned.
Increasing the number of robots yielded the opposite effects. 
RC's performance in the second setting was consistently weak
such that little difference was observed as more tasks or robots were added.

Finally, we tested all three methods in both settings with varying number of CIRs.
Increasing the number of CIRs worsened the relative performance of STAMR-G initially and,
unexpectedly, \textit{improved} it as the number continued to climb. 
As the number of CIRs increased, the restrictive aspects played a more substantial role, limiting multitasking such that RC approached STAMR and STAMR-G.
%With even higher number of CIRs that are enough to probabilistically prevent each robot from taking more than one task, RC would likely catch up with STAMR and STAMR-G.
% {\color{black}Note that in all cases, computation for STAMR and STAMR-G took significantly longer than RC.
% This is expected; STAMR and STAMR-G were both solving a more difficult problem with multitasking enabled: they must reason about multitasking feasibility and then optimize.}
%preliminary solutions to demonstrate the benefits of multitasking. %being essentially brute-force.
%Future work will focus on finding a more efficient solution.

\subsection{Site Clearing}

%In this evaluation, 
{We used STAMR with simulated robots} in site-clearing scenarios 
where we solved %used four simulated robots to solve 
a more complex version of our running example (Fig. \ref{fig:experiment}).  %, shown in Fig. \ref{fig:experiment}.
%In our experiment, 
We used a green line at the top %on the ground 
to denote a ``point of no return'' for the robots: they may each only cross the line once.
Shown in Fig. \ref{fig:experiment} are scenarios involving four robots %``waffle'' Turtlebots
and six boxes (except for the last).
The box weights (black) and robot pushing powers (red) are labeled in the figure.
We assume that pushing powers are additive.
% The purple boxes are heavier than the orange ones, while the green box is the heaviest.
% The two leftmost robots can push with enough force to move two %with medium strength, enough to push one or two
% of the purple %darker small 
% boxes stacked together.
% The two rightmost robots can push more strongly, and together they can push the large green box without any purple
% boxes on top of it.
% The orange boxes have the least weight. %can be added on to any stack; their weights are negligible.
Every box is associated with its own task and gives the same utility. % requiring the box to be moved.
The problem is more formally specified as follows for the first scenario (leftmost) in Fig. \ref{fig:experiment}:
% {\color{red} Assuming the following is for the first scenario, you have two robots that can push and two that push hard. This is inconsistent with the pushing powers shown in the figure. Also, $o_4$ should be $weight8$ now below?}
% {\color{purple} Fixed; thanks.}

\begin{itemize}
\vskip-5pt
    \itemq contains CIRs to simulate addition of weights down stacks of boxes. For example: $\{F_{On}(X, Y), F_{Weight1}(X), F_{Weight2}(Y)\}\rightarrow F_{Weight3}(Y)$: stacking a box of weight $1$ on top of a box of weight $2$ results in an effective weight of $3$ for the bottom box. Each combination of two weights totaling less than or equal to 16 is simulated in this way.
    Similarly, $Q$ contains a second type of CIRs for pushing.
    For example, $\{F_{PushHard}(X, Y), F_{Push}(Z, Y), F_{Weight10}(Y)\} \\ \rightarrow F_{Moved}(Y)$:
    a strong push and a normal push together are sufficient for a box of weight $10$.
%\rightarrow F_{Moved}(Y)$, $q_3 = \{F_{Push}(X, Y), F_{Weight4}(Y)\} %\rightarrow F_{Moved}(Y)$, $q_4 = \{F_{Push}(X, Y), F_{Weight2}(Y)\} %\rightarrow F_{Moved}(Y)$, 
    Finally, $\{F_{On}(X, Y), F_{Moved}(X)\} \rightarrow F_{Moved}(Y)$, as before. 
    \itemr %$\{r_1, r_2, r_3, r_4\}$. 
    $r_1 = (C_{Push}(r_1, X))$, $r_2 = (C_{PushWeak}(r_2, X))$, $r_3 = (C_{PushHard}(r_3, X))$, $r_4 = (C_{PushWeak}(r_4, X))$
%    \itemt $t_1 = \{F_{Moved}(o_1), 1\}$, $t_2 = \{F_{Moved}(o_2), 1\}$, $t_3 = \{F_{Moved}(o_3), 1\}$, $t_4 = \{F_{Moved}(o_4), 1\}$, $t_5 = \{F_{Moved}(o_5), 1\}$, $t_6 = \{F_{Moved}(o_6), 1\}$
    %\itemi $F_{Weight2}(o_1), F_{Weight2}(o_2), F_{Weight2}(o_3), F_{Weight8}(o_4)$, $F_{Weight1}(o_5), F_{Weight1}(o_6)$
    \itemd produces candidate initial states; each scenario in Fig. \ref{fig:experiment} corresponds to a chosen candidate initial state. 
    %as possible stacks of boxes {\color{blue} from the initial state 
    % shown in the leftmost of Fig. \ref{}} {\color{purple} I'm not 100\% sure what you were going for here, so I left it unedited,}  %as shown in Fig. \ref{fig:experiment}, 
    % as appropriate for each scenario.
\end{itemize}
%\textcolor{red}{You may also want to provide more meaningful variable names above. See my comment in Order Delivery section below. It is very difficult to interpret these rules without meaningful names. Like R corresponds to any robot, and B corresponds to any box, etc.}

Assume all the boxes are on the ground initially.
Under the box weights and robot pushing powers shown in the first scenario (leftmost figure) in Fig. \ref{fig:experiment}, one of the optimal solutions %The optimal solution, 
%shown in the same figure, 
involves stacking the boxes as shown in the same figure before task allocation (via $\Delta$). 
%purple box from the green box, stack two of the purple boxes together so that the robots on the left can clear them, 
% and stack the orange boxes onto the large green box for the two rightmost robots to push together.
% {$\delta$* handles the }unstacking and stacking - updating $I$ to induce multitasking opportunities - before task assignments.
%This evaluation illustrates a scenario that involves complex interactions involving multitasking robot and multi-robot tasks. 
%STAMR and STAMR-G were both able to optimally solve this scenario.
Next, we show that 1) our approach can generate various solutions under different candidate initial states to show its flexibility, and 2) our approach can generalize to novel scenarios with changes to the original problem.
% We also examined the effects of different initial configurations (I and I') on task assignments.
We first examined the effect of a different candidate initial state (second scenario).
Our approach was flexible enough to generate both solutions even though they were quite different (i.e., one involving multi-robot tasks and the other one with only single-robot tasks).
In the third scenario, we changed the pushing powers of the robots, resulting in a situation where the optimal solution had a single robot push all the boxes. 
In the fourth and last scenario, we added more boxes than that could be pushed and our approach still had no problems. 
% Then, we gave one robot a much more powerful - but more expensive - pushing power.
% Finally, we tested the assignment algorithm with both the increased power on one robot and additional boxes for the other robots to push.
In all cases, both STAMR and STAMR-G were able to find an optimal solution without changes other than modifications to the problem described.
In every case, STAMR took  
%large 
significantly more time than other methods; this is expected and STAMR should only be used when computation time 
%- even on a large scale - 
is not a 
%relevant 
concern.

\subsection{Order Delivery}
%{\color{green}{updated product $\rightarrow$ order}}
We additionally performed an evaluation to examine both the scalability of our approach and to demonstrate its applicability.
%We use it to illustrate how our approach can handle a complicated and realistic scenario.
In this scenario, robots are to deliver orders of boxes from stores to houses (Fig. \ref{fig:experiment2}). 
While a robot can deliver a single order quickly,
delivering more than one order at a time (but 
no more than two
%less than three 
for simplicity) can be more efficient but requires the robots to move slower, due to stacking one order on another, to avoid dropping the orders. 
Too many robots delivering to the same house could contribute to congestion around that house and
slowness could aggravate congestion, which can lead to safety issues (i.e., collisions). 

We created a virtual environment in the Webots simulator \cite{webots} where %boxes must be delivered from stores to houses.
the stores are indicated by red zones, and the houses by green zones.
Local obstacle avoidance was implemented for the robots to coordinate with each other. Order pickup and delivery were simulated: when a robot got close to the store or house, orders would be automatically loaded/unloaded.
%Due to the complex constraints present, 
A simple baseline would be for each robot to deliver one order at a time. 
Next, we show how our approach handled such scenarios more flexibly.
First, we use a rule to express that robots need to run slower with more than one product:

% The robots can each move either quickly or slowly.
% The only difference between these actions is that the quicker movement has lower cost and a prerequisite, namely $\neg F_{RequireSlow}(r_i)$ where $r_i$ refers to the robot itself.
% Each robot can take one or two boxes, but taking two boxes will force the robot to move slowly, to avoid tipping over the stack.

\begin{center}
\vskip-5pt
    $F_{On}(X,Y), F_{On}(Y,R) \rightarrow F_{RequireSlow}(R)$
\end{center}
\vskip-5pt

%Three boxes is too heavy for any robot to move.
To ensure delivering 
no more than two
%fewer than three 
orders at a time, 
we introduce a CIR to create a logically invalid state on a special predicate, which the assignment algorithm recognizes as signaling an invalid assignment:

\begin{center}
\vskip-5pt
    % $F_{On}(W,X), F_{On}(X,Y), F_{On}(Y,Z) \rightarrow F_{Invalid}(Z) (\neg F_{Invalid}(Z)$)$
    % $F_{On}(W,X), F_{On}(X,Y), F_{On}(Y,Z) \rightarrow \neg F_{Invalid}(Z)$
    $F_{On}(X,Y), F_{On}(Y,Z), F_{On}(Z,R) \rightarrow \bot$
%    {\color{purple} W-MAXSAT solvers don't recognize $\bot$; do we need to explain how this will be converted?}
% no need. 
\end{center}
\vskip-5pt

Such an invalid state is recognized automatically because of the SAT conversion. These CIRs will become implication clauses such that the assignment algorithm will avoid setting the antecedent (left side) to true.  %since it is impossible to satisfy both clauses if the antecedent is fulfilled.
%and doing nothing at all (to avoid triggering the rules) will give a higher score in the SAT problem than satisfying the left side of these rules.
%Thus, the assignment algorithm will always avoid triggering the rules.
There are CIRs to calculate congestion; robots moving slowly cause more congestion at a location and robots moving quickly cause less:

\begin{center}
\vskip-5pt
    $F_{MovingQuickly}(R,L) \rightarrow F_{Congested1}(L)$
    $F_{MovingSlowly}(R,L) \rightarrow F_{Congested2}(L)$
\end{center}
\vskip-5pt
% {\color{orange}{What are the X, Y, Z in the CIRs? Confusing. I would suggest to use more meaningful variable names, such as O (Order), L (Location, which can be either a house or store), R (Robot).}}
% {\color{blue} While I respect that O/L/R are more specific to the things being referred to, I worry that readers might wonder why we're using these designations all of a sudden instead of the X/Y we've been very consistently using up to this point.}
% {\color{red}{there are variable names... If renaming them helps understanding, I don't see why we have to keep them "consistent" to make it harder to understand.}}

There are CIRs to add up congestion levels to a store or house, similar to the weight CIRs in Site Clearing.
Finally, there is a CIR for safety and efficiency by preventing excessive congestion at a location:

\begin{center}
\vskip-5pt
    % $F_{Congested4}(X) \rightarrow F_{Invalid}(X)$
    % $F_{Congested4}(X) \rightarrow \neg F_{Invalid}(X)$
    $F_{Congested4}(L) \rightarrow \bot$
\end{center}
\vskip-5pt
% {\color{orange}{Assuming X above means a location: if each house only placed 2 orders, it would never reach congested level 4, right? If so, why is this rule above even necessary? Am I missing something here?}} {\color{blue} Yes; note that MovingSlowly adds 2 congestion.}
% {\color{red}so you meant that the only case where this can occur is when a robot carries 1 order for house A and 1 order for a different house, while another robot also carries 1 order for house A and 1 order for another house?}
% {\color{blue} Or any other case where two robots, each carrying to orders, deliver to the same house.}
where we prohibit a congestion level above $4$, which is a controllable variable. 
%Without considering these rules, a naive approach would assign robots to move quickly while carrying two boxes each, and could assign robots in a way that creates unsafe congestion.
%Our approach avoids both issues, assigning robots to safely carry two boxes each and avoiding excessive congestion.
Our approach can strike a balance between efficiency and safety by avoiding excessive congestion.%, whether due to slow or fast running robots.

% {\color{orange}{How many robots? How many houses? How many stores? These need to be stated. Why is delivering more than one order at a time more beneficial (than delivering individually)? This is unclear.}}
We randomly generated the starting positions of the robots, stores, and houses with regional constraints to ensure a roughly even distribution within their designated areas.
Each order
%box
was pseudo-randomly assigned both a starting store and a destination house while ensuring that each store received 4 orders and each house placed 2 orders. %to be delivered to.
% {\color{blue} We included 5 stores, 10 houses, and 20 robots.}
We generated a single scenario {\color{black}with 5 stores, 10 houses, and 20 robots}, and ran it 10 times. 
In these scenarios, each order corresponds to a task and all tasks share the same utility. 
A robot moving quickly is associated with a lower cost (but it can deliver only one order)
while moving slowly is associated with a slightly higher cost (while delivering two).

%{\color{blue} Add in the utility and cost. 
%with a hypothesis that carrying multiple orders simultaneously would allow for greater efficiency despite the increased congestion.

%{\color{red}How is such "efficiency" evaluated in your system? Presumably, you system would need to prefer an allocation solution where two orders are delivered at a time vs another solution where orders are delivered individually. How is such a preference even recognized in your system in terms of utility and cost?}

Our approach correctly recognized that the superior solution was to have most of the robots deliver two orders to the same houses 
%(when available) 
and move slowly.
Note that due to our focus on IA (instantaneous assignment),
the allocation solution is based on some {\it conservative} assumptions, such as 
a robot that is assigned to deliver two orders always moves slowly even though in practice the robot can move faster after delivering an order; 
similarly, two robots delivering to the same house are assumed to increase congestion even when the robots may arrive at different times.
% {\color{orange}{Confused here. 1) What if the two orders are at different stores? Or that is not going to happen in the initialization? If so, it needs to be stated. 2) Also, how did you assign the utility for each order and cost? I am not sure why "most of robots delivering two orders" would be a better solution without these utilities/costs.}}
% {\color{orange}{3) What happens after an order is delivered with a robot delivering two orders at a time? Is the robot allowed to move faster then or it will always move slower. Either case, it needs to be stated.}},
% {\color{blue} If the orders are at different stores, the robot will pick up the closest one first, then move on to picking up the next one. Robots only move slower while actually carrying two products, and optimality is measured in terms of time taken, though collisions between robots are also a relevant factor to consider.}
% {\color{red}Ok. Let me see if I understand. Since your approach is IA, does it mean that you are {\bf{pessimistically}} considering that delivering to the same house can lead to a congestion and delivering 2 orders will slow things down, even though in actually, it may not even create such a congestion (i.e., robots arriving at the same house at different times) or the slowing down may only be for a part of the execution?}
% {\color{blue} Exactly so.}
%and have the rest of the orders delivered by fast-moving robots.

Because we used E-puck robots that lack strong mobility, collisions are nearly unavoidable in complex situations such as this.
However, our approach resulted in a significantly lower average number of collisions, as well as a lower average time-to-completion, indicative of fewer avoidance maneuvers.
The baseline averaged 10 collisions ($\sigma \approx 0.5$) between the 20 robots used in every run, while ours yielded an average of 1 collision ($\sigma \approx 0.08$) among 14 average robots used.
Additionally, our approach completed the deliveries approximately 24\% faster ($\sigma \approx 3\%$) on average. 

\section{Conclusions}
\vskip-3pt
In this paper, we presented a novel framework for task allocation with multitasking robots, considering both the synergistic and restrictive aspects of multitasking. % multi-robotic systems.
It addresses a gap in existing methods by
considering physical constraints as the key to enable multitasking in multi-robot systems. 
%and assignments as interdependent, thus providing a more general solution. 
%of formulating the multi-robot task assignment problem.
We formulated the problem %referred to as TAMPiC, a
and presented a compilation method to convert it to Weighted MAX-SAT.
%We showed that the conversion was both sound and complete.
We further presented a greedy method. % based on the compiled problem. 
We compared both methods with a modern ST-MR-IA baseline to demonstrate the benefits of multitasking in synthetic domains and with two simulations to demonstrate its scalability, flexibility, and applicability.
%that this new class of problem is difficult for existing algorithms to solve.

One major limitation of our work is %include a lack of an "is-a" relationship other than the hardcoded "everything is-an object" relationship and 
an absence of examination into true numeric constraints, including temporality.
Notably, addressing this would enable us to simplify the specification of CIRs substantially and  consider TA task assignment,  instead of being limited to IA, and we plan to do so in future work.
Considering numericity in constraints would remove the need for, i.e., the separate "Push" and "StrongPush" capabilities in our running example.
An extension into temporality would require a consideration of required concurrency \cite{cushing-temporality} as it applies to CIRs as well as to actions, but would significantly expand the expressiveness of our approach.
% Another interesting extension to our work would be an examination of TA assignments, including durative constraints.
% % Finally, it would be convenient to automate the reconfiguration function $\Delta$, potentially by marrying this problem with planning.
We also plan to improve the accuracy of the greedy approximation method and provide bounds on the quality of the approximation.
Finally, the specification of CIRs is not wholly intuitive; it would be a worthy extension to our work simply to create a formulation which is easier to intuit, perhaps via learning from examples.
These extensions 
%to this work 
would allow for simulation of complex environments, such as a construction domain with complex physical constraints.
% For example, we could easily simulate a problem where robots are collaboratively constructing a building, which is a setting that would be prohibitively complex to specify using CIRs, even if temporality were to be considered.

% Additionally, while %stochastic domains and imperfect communication between robots 
% addressing stochastic domains and distributed systems
% are out of the scope of this paper, these would be both interesting and relevant issues to address.

%\section*{ACKNOWLEDGMENT}

\noindent{\textbf{Acknowledgment}}: We thank the anonymous reviewers for their helpful comments and suggestions. 
This research is supported in part by NSF grant 2047186. 
 
%\section*{Acknowledgments}
%We thank the anonymous reviewers for their helpful comments. 
%This research is supported in part by the NSF grant IIS-1844524, the NASA grant NNX17AD06G, and the AFOSR grant FA9550-18- 1-0067.

\bibliographystyle{IEEEtran}
\bibliography{references}

\end{document}